\documentclass{article}
\usepackage{amssymb}
\usepackage{amsmath}
\usepackage{stmaryrd}
\usepackage{graphics}
\usepackage{tikz}
\usepackage{epsfig}
\usepackage{multirow}
\usepackage{graphicx}
\usepackage[center]{caption}
\usepackage{subcaption}
\usepackage{cancel}
\usepackage{hyperref}
\newcommand\drawbull[1][black]{%
\begin{tikzpicture}
	\draw[#1,fill=#1] (0,0) circle (.5ex); 
\end{tikzpicture}%
}
\newcommand\drawlbulls[1][black]{%
\begin{tikzpicture}
	\draw[#1!20!white,fill=#1!20!white] (0,0) circle (.5ex);
	\draw[#1!23!white,fill=#1!23!white] (0.1,0) circle (.5ex);
	\draw[#1!26!white,fill=#1!26!white] (0.2,0) circle (.5ex);
	\draw[#1!29!white,fill=#1!29!white] (0.3,0) circle (.5ex);
	\draw[#1!32!white,fill=#1!32!white] (0.4,0) circle (.5ex);
\end{tikzpicture}%
}

\newcommand\drawdbulls[1][black]{%
\begin{tikzpicture}
	\draw[#1!40!white,fill=#1!40!white] (0,0) circle (.5ex);
	\draw[#1!43!white,fill=#1!43!white] (0.1,0) circle (.5ex);
	\draw[#1!46!white,fill=#1!46!white] (0.2,0) circle (.5ex);
	\draw[#1!49!white,fill=#1!49!white] (0.3,0) circle (.5ex);
	\draw[#1!52!white,fill=#1!52!white] (0.4,0) circle (.5ex);
\end{tikzpicture}%
}
\newcommand\drawbullsmooth[1][black]{%
\begin{tikzpicture}
	\draw[#1!25!white,fill=#1!25!white] (0,0) circle (.5ex);
	\draw[#1!33!white,fill=#1!33!white] (0.1,0) circle (.5ex);
	\draw[#1!42!white,fill=#1!42!white] (0.2,0) circle (.5ex);
	\draw[#1!48!white,fill=#1!48!white] (0.3,0) circle (.5ex);
	\draw[#1,fill=#1] (0.4,0) circle (.5ex);
\end{tikzpicture}%
}
\newcommand\drawbullosc[1][black]{%
\begin{tikzpicture}
	\draw[#1!30!white,fill=#1!30!white] (0,0) circle (.5ex);
	\draw[#1!36!white,fill=#1!36!white] (0.1,0) circle (.5ex);
	\draw[#1!42!white,fill=#1!42!white] (0.3,0) circle (.5ex);
	\draw[#1!48!white,fill=#1!48!white] (0.4,0) circle (.5ex);
	\draw[#1,fill=#1] (0.2,0) circle (.5ex);
\end{tikzpicture}%
}

\title{Real-valued continued fraction of straight lines}
\author{Vijay Prakash S \\
        \small Alappuzha, Kerala, India. \\
        \small \tt{prakash.vijay.s@gmail.com} \\
}
\date{}
\begin{document} 
\maketitle
\begin{abstract}
\noindent In an unbounded plane, straight lines are used extensively for mathematical analysis. They are tools of convenience. However, those with high slope values become unbounded at a faster rate than the independent variable. So, straight lines, in this work, are made to be bounded by introducing a parametric nonlinear term that is positive. The straight lines are transformed into bounded nonlinear curves that become unbounded at a much slower rate than the independent variable. This transforming equation can be expressed as a continued fraction of straight lines. The continued fraction is real-valued and converges to the solutions of the transforming equation. Following Euler's method, the continued fraction has been reduced into an infinite series. The usefulness of the bounding nature of continued fraction is demonstrated by solving the problem of image classification. Parameters estimated on the Fashion-MNIST dataset of greyscale images using continued fraction of regression lines have less variance, converge quickly and are more accurate than the linear counterpart. Moreover, this multi-dimensional parametric estimation problem can be expressed on $xy-$ plane using the parameters of the continued fraction and patterns emerge on planar plots. 
\end{abstract}
\section{Introduction}
\begin{figure}[!htbp]
\centerline{\includegraphics[width=0.5\textwidth]{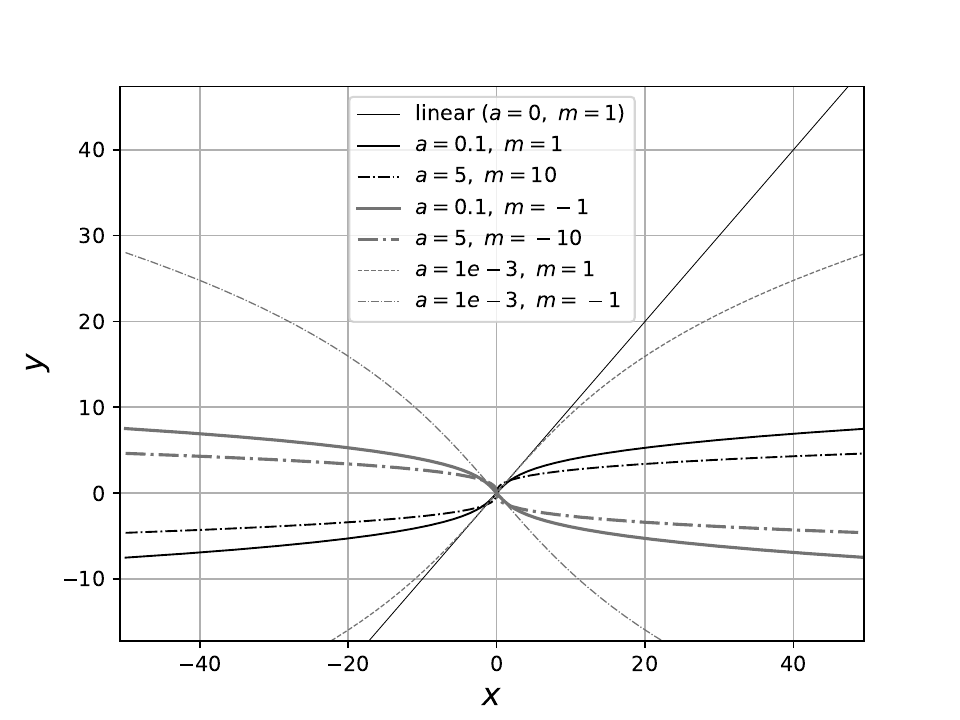}}
\caption{$y$ for different $a$ and $m$ values. As $a\rightarrow\infty$ we approach the $x-$ axis.}
\label{fig1}
\end{figure}

Nonlinearity is often expressed as various powers of the independent variable $x$ on the $xy-$ plane. Such expressions, especially those with high powers of $x$, make $y$ sensitive to small changes in $x$ and become unbounded. Perhaps, the most convenient equation to plot on an $xy-$plane is $y=x$, which can be parameterized as $y=mx$. Any point on the $xy-$plane can be expressed with $y=mx$, which has $y-$axis at the limit $m\rightarrow\pm\infty$. However, even for $m>1$, the dependent $y-$values become greater than the independent $x-$values and hence can become unbounded for large values of $x$. So, even a straight line expression can lead to unbounded situations given $m\in(-\infty,\infty)$.
To keep the dependent values bounded, we introduce a nonlinear term as follows
\begin{equation}
(1+ay^2)y=mx.
\label{eq1}
\end{equation}
In the above equation, $a>0$ and the term $ay^2>0$. So, $y$ values are always less than the $x$ values. Eqn. (\ref{eq1}) is plotted in Fig. (\ref{fig1}) for various $a$ and $m$. 
It can be seen from Fig. (\ref{fig1}) that $y$ of Eqn. (\ref{eq1}) is always bounded by $y=mx$ lines.

\begin{figure}[ht!]
\centering
 \begin{subfigure}[b]{0.45\textwidth}
\includegraphics[width=\textwidth]{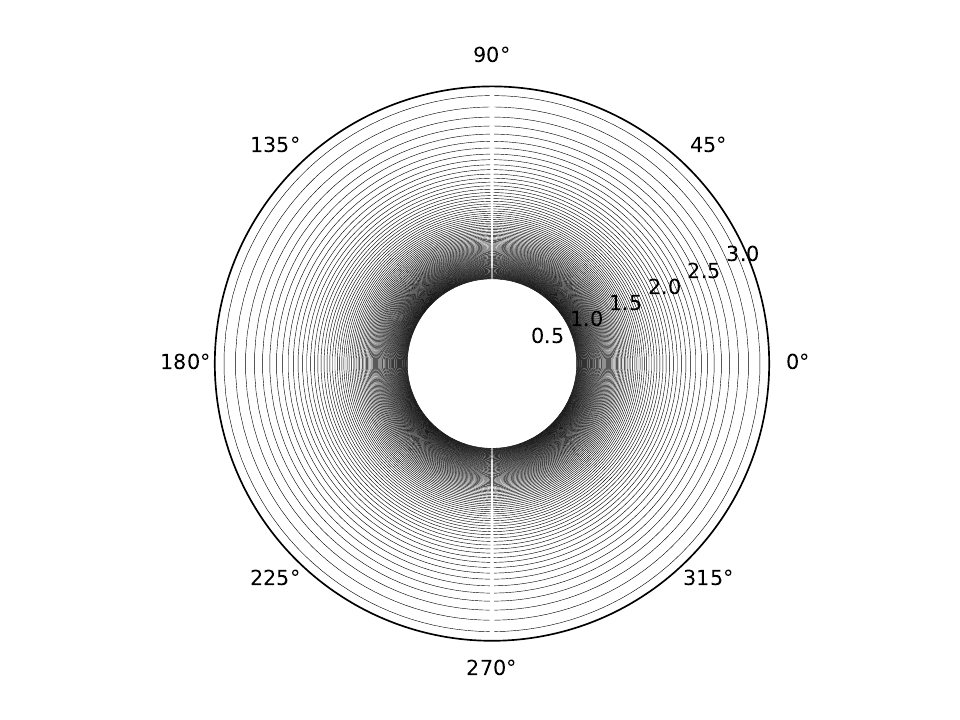}
\caption{$a\in [1e-1,1]$  and $m\in$ interval [-200,200].}
 \label{fig2a}
\end{subfigure}
\begin{subfigure}[b]{0.45\textwidth}
\includegraphics[width=\textwidth]{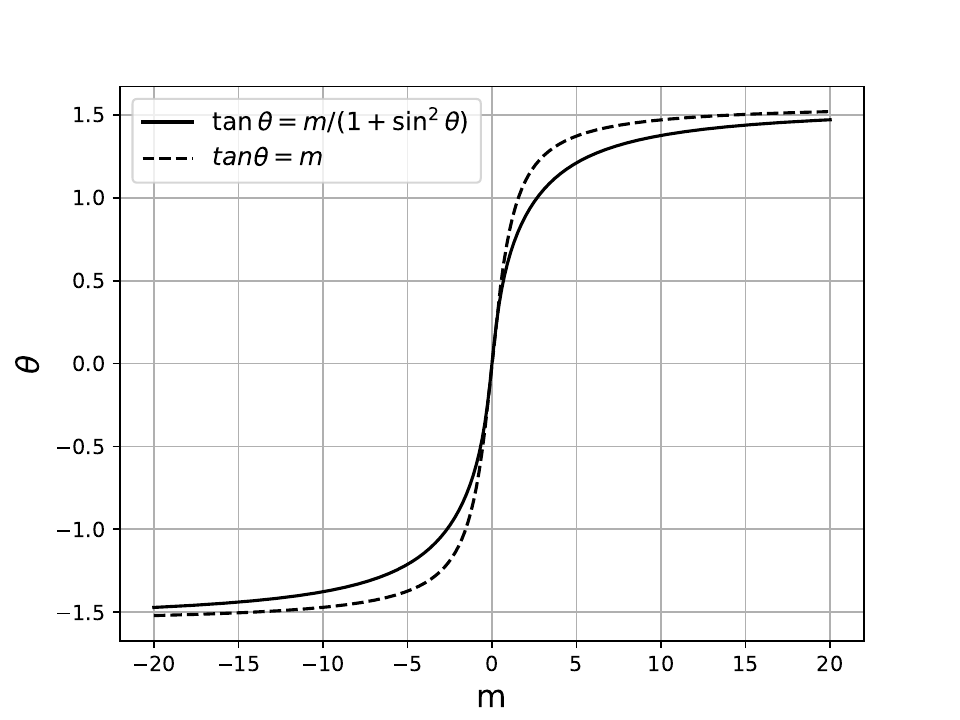}
\caption{Reduced phase when compared to the linear case.}
 \label{fig2b}
\end{subfigure}
\caption{Polar plots of Eqn. (\ref{eq1}) with $r=1/\sqrt{a}$}
\label{fig2}
\end{figure}

In polar co-ordinates $(r,\theta)$, for $x=r\cos\theta$ and $y=r\sin\theta$, Eqn. (\ref{eq1}) becomes
\begin{equation}
\nonumber\tan\theta(1+ar^2\sin^2\theta) = m.
\end{equation}
Choosing $r=1/\sqrt{a}$, we get
\begin{equation}
\tan\theta(1+\sin^2\theta) = m.
\label{eq2}
\end{equation}
Solving the above equation (Eqn. (\ref{eq2})) numerically, we get the polar plots as shown in  Fig. (\ref{fig2}). The curves reach origin in the asymptotic limit $a\rightarrow\infty$.

In the following section, we will discuss about the real solution of Eqn. (\ref{eq1}) and its properties. In Section 3, in order to demonstrate the bounding nature of Eqn. (\ref{eq1}), we consider the problem of classification of greyscale images of the Fashion-MNIST dataset into 10 categories.  Discussions on the results of classification are provided in Section 4. Finally, we conclude our work in Section 5.
\section{Real solution and its properties}
Eqn. (\ref{eq1}) has two complex roots and a real root for all $a>0$. The real solution is given by
\begin{eqnarray}
y = -\frac{1}{3}{t^\frac{1}{3}}+\frac{1}{a}{t^{-\frac{1}{3}}}, \textnormal{where }
t=-\left(\frac{27mx}{2a}\right)+\sqrt{\left(\frac{27mx}{2a}\right)^2+\frac{27}{a^3}}.
\label{eq3}
\end{eqnarray}
The solution of $y$ is sum of two components  i: $-\frac{1}{3}{t^\frac{1}{3}}$ and ii: $\frac{1}{a}{t^{-\frac{1}{3}}}$. 
\begin{figure}[ht!]
\centering
 \begin{subfigure}[b]{0.46\textwidth}
\includegraphics[width=\textwidth]{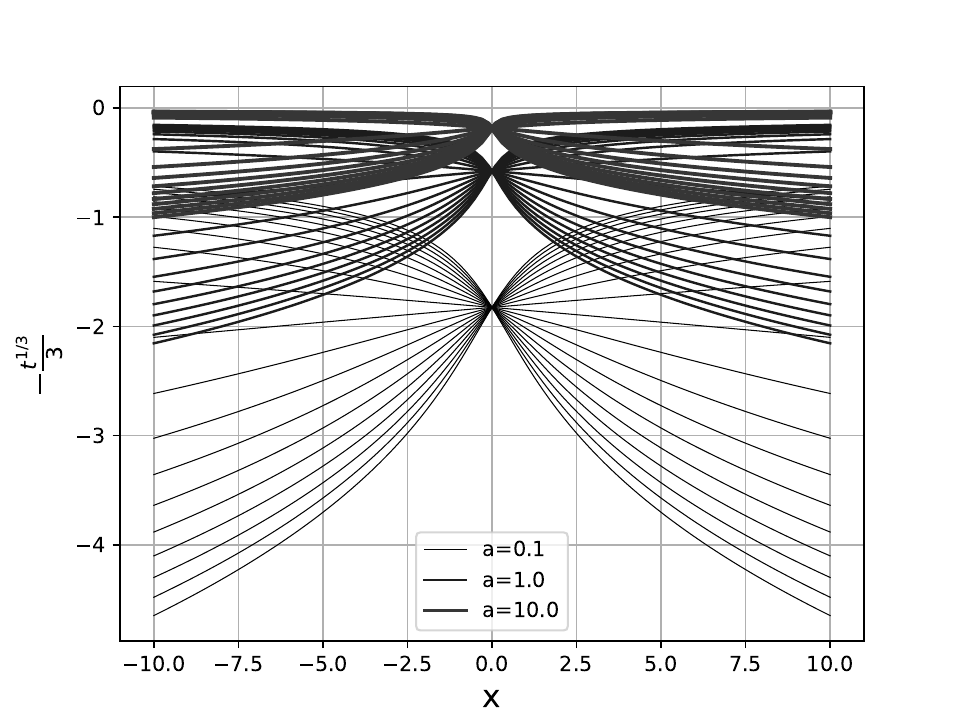}
\caption{Plots of i: $-\frac{1}{3}{t^\frac{1}{3}}$ for various $a$ and $m\in$ interval [-1,1].}
 \label{fig3a}
\end{subfigure}
\begin{subfigure}[b]{0.46\textwidth}
\includegraphics[width=\textwidth]{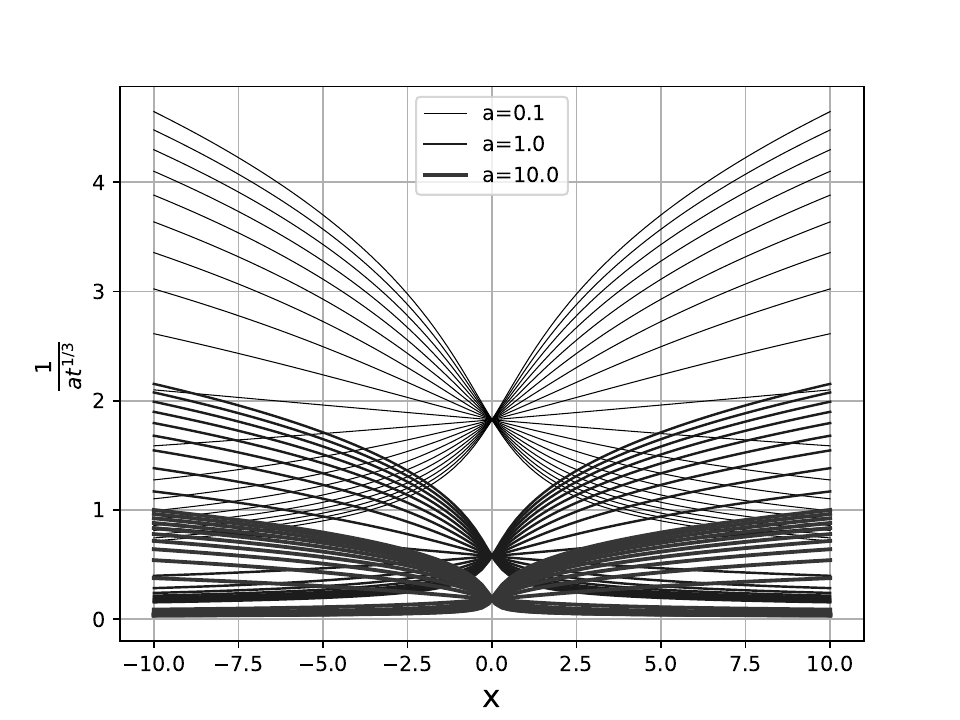}
\caption{Plots of ii: $\frac{1}{a}{t^{-\frac{1}{3}}}$ for various $a$ and $m\in$ interval [-1,1].}
 \label{fig3b}
\end{subfigure}
\caption{Plots of individual components i and ii of $y$ of Eqn. (\ref{eq3}).}
\label{fig3}
\end{figure}
The plots of these two components are shown in Fig. (\ref{fig3}) for various values of $a$ and $m$. The two components i and ii do not intersect on the $xy-$ plane for various values of $a$ and $m$. This is also shown in Fig. \ref{fig4a}. Since i and ii do not intersect we can consider them as axes and obtain plots for various $a$ and $m$ values. This is shown in Fig. \ref{fig4b}. The origin of i,ii- plane is at the limit $a\rightarrow\infty$, as i$\rightarrow0$ and ii$\rightarrow0$.

\begin{figure}[ht!]
\centering
 \begin{subfigure}[b]{0.46\textwidth}
\includegraphics[width=\textwidth]{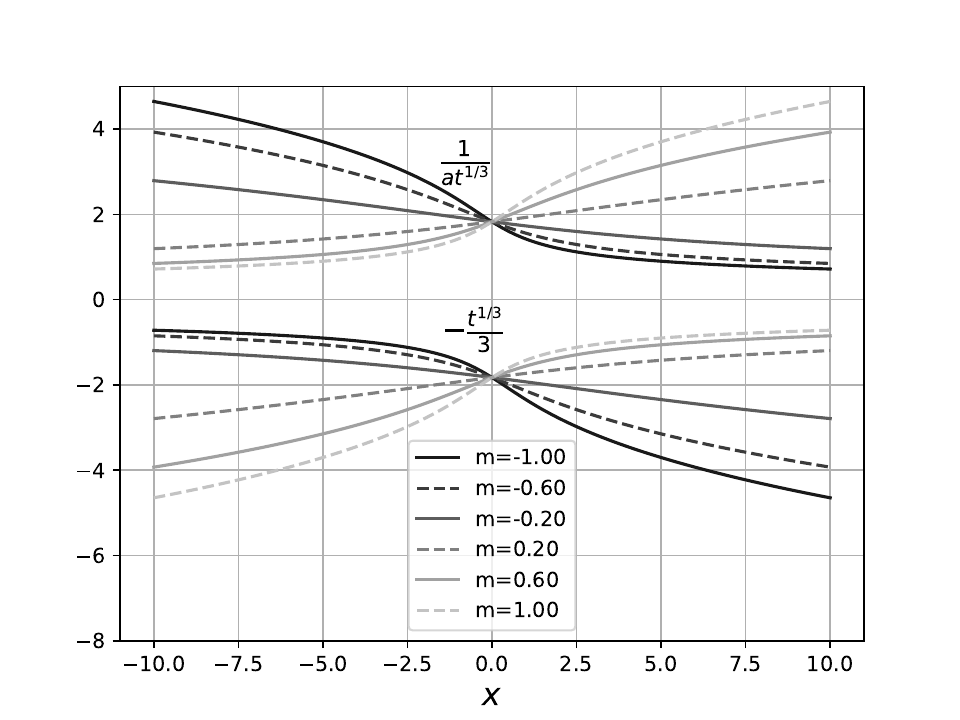}
\caption{Plots of i and ii for $a=0.1$ and for various $m\in[-1,-0.6,-0.2,0.2,0.6,1]$.}
 \label{fig4a}
\end{subfigure}
\begin{subfigure}[b]{0.46\textwidth}
\includegraphics[width=\textwidth]{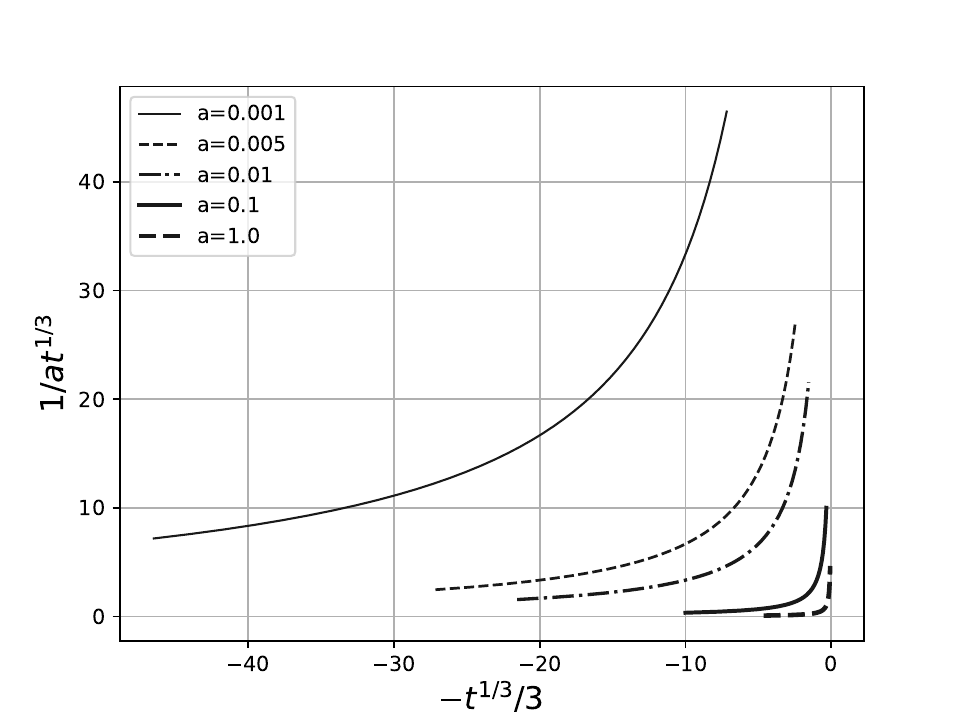}
\caption{Plots of various $a$ in (i,ii)- plane with $m\in$ interval [-100,100].}
 \label{fig4b}
\end{subfigure}
\caption{(a) shows i and ii for a fixed $a$. (b) With i and ii as axes, we get unique curves for each $a\in[1e-3,5e-3,1e-2,0.1,1]$ and for various values of $m$.}
\label{fig4}
\end{figure}
\paragraph{Properties:}

\begin{enumerate}
\item As mentioned in the previous section, as $a\rightarrow\infty$ the i+ii curves approach $x-$axis on the $xy-$ plane.
\item From Fig. (\ref{fig4b}) with components i and ii as axes, it can be seen that we only have to determine $a$ in order to obtain unique solutions on the (i,ii)- plane. So, in Section 3, we estimate parameters of the problem based on the estimate of $a$. Once $a$ converges, the other parameters converge.
\item Eqn. (\ref{eq1}) can be written as
\begin{equation}
\nonumber y = \frac{mx}{1+ay^2},
\end{equation}
which can be expressed as a continued fraction
\begin{eqnarray}
\nonumber y &=& \frac{mx}{1+a\frac{m^2x^2}{\left( 1+a\left( \frac{mx}{1+\cdots}\right)^2\right)^2}}.
\end{eqnarray}
Let us re-write the above equation as
\begin{equation}
\nonumber \hat{y} = \frac{1}{1+\hat{a}\frac{1}{\left( 1+\hat{a}\left( \frac{1}{1+\cdots}\right)^2\right)^2}},
\end{equation}
where $\hat{y}=y/(mx)$ and $\hat{a}=am^2x^2$. Following Euler's method \cite{euler}, we will find simple fractions that continuously approach $\hat{y}$. On the right-hand side, initially we have $1/(1+\hat{a})$. We have the numerator $A=1$ and the denominator $\mathcal{A}= 1+\hat{a}$.
\begin{figure}[!htbp]
\centerline{\includegraphics[width=0.6\textwidth]{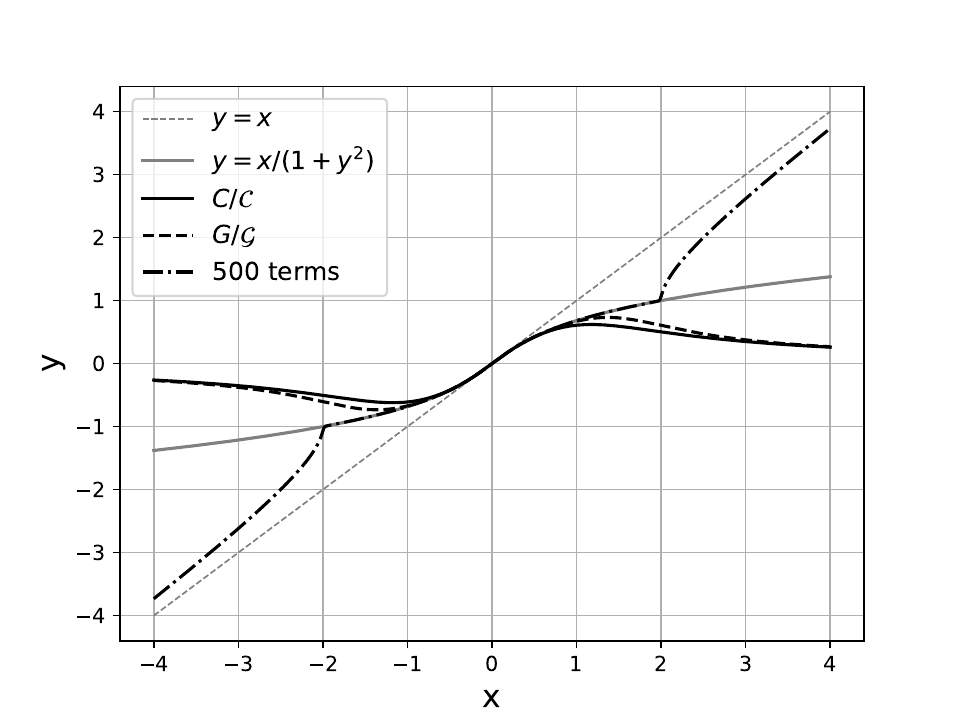}}
\caption{Continued fraction series converging to $y=1/at^{1/3}-t^{1/3}/3$ or $y=\text{i}+\text{ii}$ values. All the curves including the truncated Eqn. (\ref{eqseries1})  are bounded by $y=x$ line. $a=1,\ m=1$}
\label{fig5}
\end{figure}
Let us reduce the continued fraction into an infinite series \cite{euler} with the first fraction being $A/\mathcal{A}$. Further we have,
\begin{eqnarray}
\nonumber\frac{B}{\mathcal{B}}=\frac{1}{1+\hat{a}\frac{1}{\left( 1+\hat{a}\right)^2}}
&=& \frac{( 1+\hat{a})^2}{( 1+\hat{a})^2+a} = \frac{\mathcal{A}^2}{\mathcal{A}^2+\hat{a}},\\
\nonumber\frac{C}{\mathcal{C}}=\frac{1}{1+\hat{a}\frac{1}{\left( 1+\hat{a}\left(\frac{1}{1+\hat{a}}\right)^2\right)^2}}&=&\frac{\mathcal{B}^2}{\mathcal{B}^2+\hat{a}B^2},\\
\nonumber\frac{D}{\mathcal{D}}=\frac{1}{1+\hat{a}\frac{1}{\left( 1+\hat{a}\left(\frac{1}{1+\hat{a}\left(\frac{1}{1+\hat{a}}\right)^2}\right)^2\right)^2}} &=& \frac{\mathcal{C}^2}{\mathcal{C}^2+\hat{a}C^2},
\end{eqnarray}
and so on. The last of the above fractions is closer to the $\hat{y}$ value. Now,
\begin{eqnarray}
\nonumber\frac{B}{\mathcal{B}}-\frac{A}{\mathcal{A}}&=& \frac{\mathcal{A}^2}{\mathcal{A}^2+\hat{a}}-\frac{1}{\mathcal{A}} = \frac{\mathcal{A}^3-(\mathcal{A}^2+\hat{a})}{\mathcal{A}(\mathcal{A}^2+\hat{a})}\\
\nonumber\Rightarrow \frac{B}{\mathcal{B}}&=&\frac{1}{1+\hat{a}}+\frac{\mathcal{A}^3-(\mathcal{A}^2+\hat{a})}{\mathcal{A}(\mathcal{A}^2+\hat{a})},\\
\nonumber \frac{C}{\mathcal{C}}-\frac{B}{\mathcal{B}}&=&\frac{\mathcal{B}^2}{\mathcal{B}^2+\hat{a}B^2}-\frac{B}{\mathcal{B}}=\frac{\mathcal{B}^3-B(\mathcal{B}^2+\hat{a}B^2)}{\mathcal{B}(\mathcal{B}^2+\hat{a}B^2)}\\
\nonumber\Rightarrow \frac{C}{\mathcal{C}} &=&\frac{1}{1+\hat{a}}+\frac{\mathcal{A}^3-(\mathcal{A}^2+\hat{a})}{\mathcal{A}(\mathcal{A}^2+\hat{a})}+\frac{\mathcal{B}^3-B(\mathcal{B}^2+\hat{a}B^2)}{\mathcal{B}(\mathcal{B}^2+\hat{a}B^2)}.\\
\nonumber\text{Similarly, }\frac{D}{\mathcal{D}}&=&\frac{1}{1+\hat{a}}+\frac{\mathcal{A}^3-(\mathcal{A}^2+\hat{a})}{\mathcal{A}(\mathcal{A}^2+\hat{a})}+\frac{\mathcal{B}^3-B(\mathcal{B}^2+\hat{a}B^2)}{\mathcal{B}(\mathcal{B}^2+\hat{a}B^2)}+\frac{\mathcal{C}^3-C(\mathcal{C}^2+\hat{a}C^2)}{\mathcal{C}(\mathcal{C}^2+\hat{a}C^2)},
\end{eqnarray}
and so on. The last term $D/\mathcal{D}$ is relatively closer to the $\hat{y}$ value. Thus, we have reduced the continued fraction into an infnite series given by
\begin{eqnarray}
\nonumber\hat{y} &=& \frac{1}{1+\hat{a}}+\frac{\mathcal{A}^3-(\mathcal{A}^2+\hat{a})}{\mathcal{A}(\mathcal{A}^2+\hat{a})}+\frac{\mathcal{B}^3-B(\mathcal{B}^2+\hat{a}B^2)}{\mathcal{B}(\mathcal{B}^2+\hat{a}B^2)}+\frac{\mathcal{C}^3-C(\mathcal{C}^2+\hat{a}C^2)}{\mathcal{C}(\mathcal{C}^2+\hat{a}C^2)}+\cdots\\
\nonumber\Rightarrow \frac{y}{mx} &=& \frac{1}{1+am^2x^2}+\frac{(1+am^2x^2)^3-((1+am^2x^2)^2+am^2x^2)}{(1+am^2x^2)((1+am^2x^2)^2+am^2x^2)}\\\nonumber
&&+\frac{((1+am^2x^2)^2+am^2x^2)^3-(1+am^2x^2)^2[((1+am^2x^2)^2+am^2x^2)^2+am^2x^2(1+am^2x^2)^4]}{((1+am^2x^2)^2+am^2x^2)((1+am^2x^2)^2+am^2x^2)^2+am^2x^2(1+am^2x^2)^4)}\\&&+\cdots
\label{eqseries1}
\end{eqnarray}
It can be seen from Fig. (\ref{fig5}) that when boundedness is introduced to $y=mx$ lines through continued fractions, $y$ converges to i+ii curves.
\item \textit{When two $y=$ i+ii curves intersect on the $xy-$plane the curves have unique $a$ and $m$ values.} 
Let us consider two intersecting curves. From Eqn. (\ref{eq1}),
\begin{eqnarray}
\nonumber y &=&\frac{m_1x}{1+a_1y^2}\ \text{and}\\
\nonumber y &=&\frac{m_2x}{1+a_2y^2}.
\end{eqnarray}
When the two curves intersect we have
\begin{eqnarray}
\nonumber \frac{m_1x}{1+a_1y^2} &=&\frac{m_2x}{1+a_2y^2}\\
\Rightarrow y^2 &=& \frac{m_2-m_1}{m_1a_2-m_2a_1}
\label{eq5}
\end{eqnarray}
When $m_1=m_2$ we get $y=0$ and when $a_1=a_2$ we get $y^2$ to be a negative quantity which is not possible in the $xy-$plane. Hence, when two i+ii curves intersect, they have unique $a$ and $m$ values. Also, from Fig. (\ref{fig4b}), we obtain unique solutions for each $a$. Hence, once we determine and fix $a$, the curves do not intersect except at $y=0$.
\end{enumerate}
With $a$ and $m$ as parameters of real-valued continued fraction of straight lines, we will consider an image classification problem in the following section. This is because the problem has sufficient sample points to estimate $a$ and $m$ and study their influence on the estimation of linear parameters of regression (Chapter 3 of \cite{method_book}). The classification is done based on the estimation of parameters of regression using the gradient descent algorithm \cite{grad_desc}. The additional parameters $a$ and $m$ play key roles in convergence and accuracy.
\section{Image classification problem}
\begin{figure}[!htbp]
\centerline{\includegraphics[width=1\textwidth]{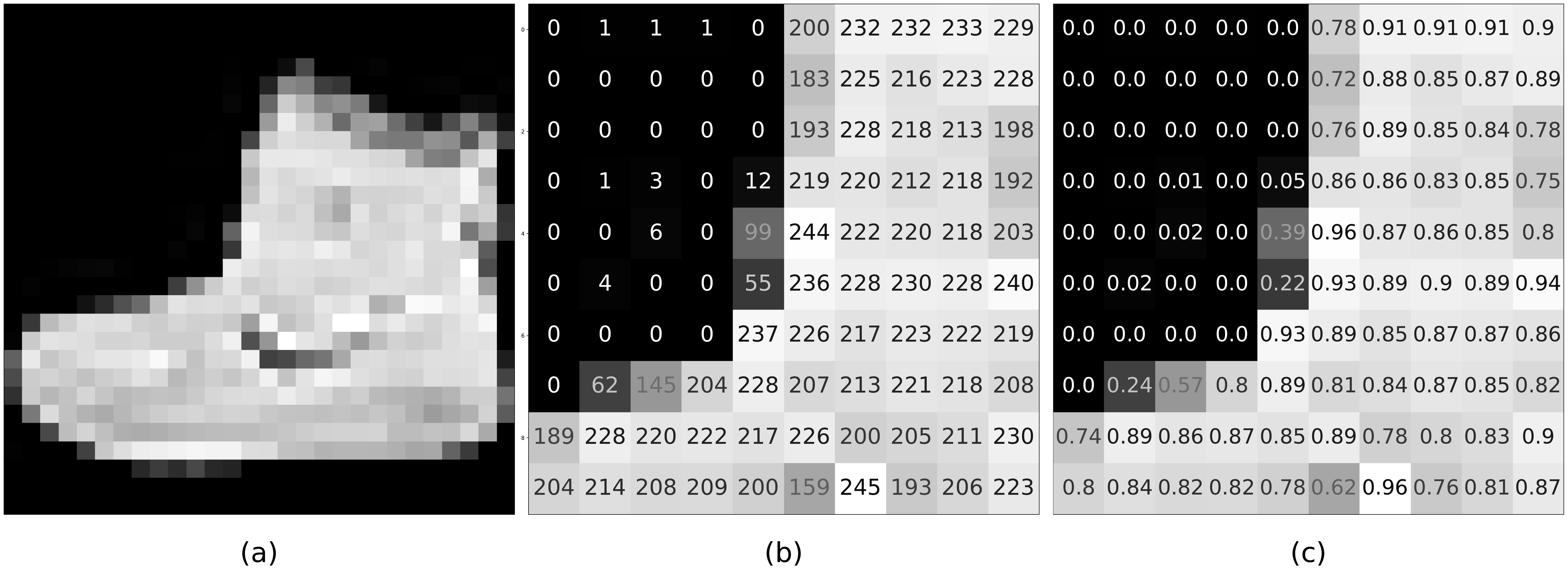}}
\caption{(a) One of the sample images of Fashion-MNIST dataset, (b) A section of the sample image with pixel values shown and (c) Normalized values of (b)}
\label{fig6}
\end{figure}

In this section, we will solve an image classification problem using the continued fraction of regression lines and compare with the results obtained using regression lines. We will classify 28x28 greyscale images of Fashion-MNIST dataset \cite{f_mnist} into 10 categories or classes by estimating the parameters that are used to find class probabilities. This dataset is more challenging in terms of achieving high accuracy of classification \cite{f_mnist}. We use gradient-descent method \cite{grad_desc} to estimate parameters including $a$ and $m$ for each category.
\paragraph{Parametric representation:}
One of the 28x28 greyscale images is shown in Fig. \ref{fig6} which belongs to one of the  10 categories present in the dataset. The dataset contains 60,000 such training samples and 10,000 test samples. 
Each pixel value is represented as a component of the input vector $\textbf{x}$. The representation is a linear combination:
\begin{eqnarray}
w_0+\sum_{i=1}^{n=784} w_ix_i = w^T\textbf{x}
\label{eqwx}
\end{eqnarray}
where $w_i'$s are the parameters of regression with $x_0=1$ included in $\textbf{x}$ for simpler notation. So that, $\textbf{x}$ has $28\times 28=784+1$ components.
Continued fraction of $w^T\textbf{x}$ converges to the real solution of Eqn. (\ref{eq1}) which is now given by
\begin{eqnarray}
ay^3+y&=&m(w^T\textbf{x})
\label{eqcfwtx}\\
\nonumber \Rightarrow y(w^T\textbf{x}) &=& -\frac{1}{3}{t^\frac{1}{3}}+\frac{1}{a}{t^{-\frac{1}{3}}}, \textnormal{where now\ }\\
t&=&-\left(\frac{27m(w^T\textbf{x})}{2a}\right)+\sqrt{\left(\frac{27m(w^T\textbf{x})}{2a}\right)^2+\frac{27}{a^3}}.
\label{eqywtx}
\end{eqnarray}
$y$ is then used to obtain class probabilities through the logistic or sigmoid function, so that we get outputs between 0 and 1. The sigmoid function is given by
\begin{equation}
\nonumber \sigma(y) = \frac{1}{1+\exp(-y)},
\end{equation}
which is compared with the logistic regression function
\begin{equation}
\nonumber \sigma(w^T\textbf{x}) = \frac{1}{1+\exp(-w^T\textbf{x})}.
\end{equation}
$\sigma(w^T\textbf{x})$ is a special case of $\sigma(y)$ with $m=1$ and $a_{\text{asymp}}=0$, where $a_{\text{asymp}}$ is $a$ at its asymptotic limit $a\rightarrow 0$.
The parameters are estimated using the provided outputs  from the 60,000 training samples. The outputs are expressed as one-hot encoded vectors, i.e., for example, if $j^{\text{th}}$ sample belongs to the $3^{\text{rd}}$ category, then the output is encoded as $p^*_{j} = [0,0,0,1,0,0,0,0,0,0]$. Therefore, $p^*_{j}$ is the optimal desired output of $\sigma$.
Using the training samples and their output vectors $p^*_{j}$'s, the parameters are chosen to maximize the joint probability of the corresponding category in $\sigma$. The parametric estimation is measured with the following loss function
\begin{equation}
L(w_0,w_1,..,w_{784},a,m)=-\frac{1}{n}\sum_{j}^{n}p^*_{j}\log \sigma_{j},
\label{eqloss}
\end{equation}
which is the negative log likelihood function over $n$ samples. The $\sigma(y)$ form of loss function and $\sigma(w^T\textbf{x})$ form of loss function are used to estimate $w_i'$s of $y(w^T\textbf{x})$ and $w_i'$s of $w^T\textbf{x}$, respectively.

The parameters are estimated using the gradient descent optimization algorithm. In order to minimize Eqn. (\ref{eqloss}), starting from initial values, the parameters $a$, $m$ and $w_i'$s of Eqn. (\ref{eqwx}) are estimated in the direction opposite to the gradient of the loss function:
\begin{eqnarray}
w_i = w_i^{\text{(prev)}} - \frac{\partial L}{\partial w_i},
m = m^{\text{(prev)}} - \frac{\partial L}{\partial m},\  \text{and }
a = a^{\text{(prev)}} - \alpha\frac{\partial L}{\partial a},
\label{equpdates}
\end{eqnarray}
where $w_i^{\text{(prev)}}$, $m^{\text{(prev)}}$ and $a^{\text{(prev)}}$ are previous values of the respective parameters and $\alpha$ is the step size. ${\partial L}/{\partial w_i}$, ${\partial L}/{\partial m}$ and ${\partial L}/{\partial a}$ are given by
\begin{eqnarray}
\nonumber\frac{\partial L}{\partial w_i} =(\sigma-p^*)\frac{\partial y}{\partial w_i},
\frac{\partial L}{\partial m} =(\sigma-p^*)\frac{\partial y}{\partial m}
\ \text{and }
\frac{\partial L}{\partial a}=(\sigma-p^*)\frac{\partial y}{\partial a},
\end{eqnarray}
respectively. The derivatives with respect to $y$ are obtained from Eqn. (\ref{eqcfwtx}) as follows:
\begin{eqnarray}
\frac{\partial y}{\partial w_i} = \frac{mx_i}{1+3ay^2},\ 
\frac{\partial y}{\partial m} = \frac{w^T\textbf{x}}{1+3ay^2},\ \text{and }
\frac{\partial y}{\partial a} = \frac{-y^3 }{1+3ay^2}.
\label{eqstep}
\end{eqnarray}
Note that usually step size is introduced for all the parameters that are estimated using the gradient descent algorithm \cite{grad_desc}. But here we do not need a step size for $m$ and $w_i'$s, since stepping is adaptive and bounded using the positive term $3ay^2$ in Eqn.  (\ref{eqstep}). The step size $\alpha$ is introduced so that the constraint $a>0$ is satisfied while stepping. 

\begin{figure}[ht!]
\centering
 \begin{subfigure}[b]{0.45\textwidth}
\includegraphics[width=\textwidth]{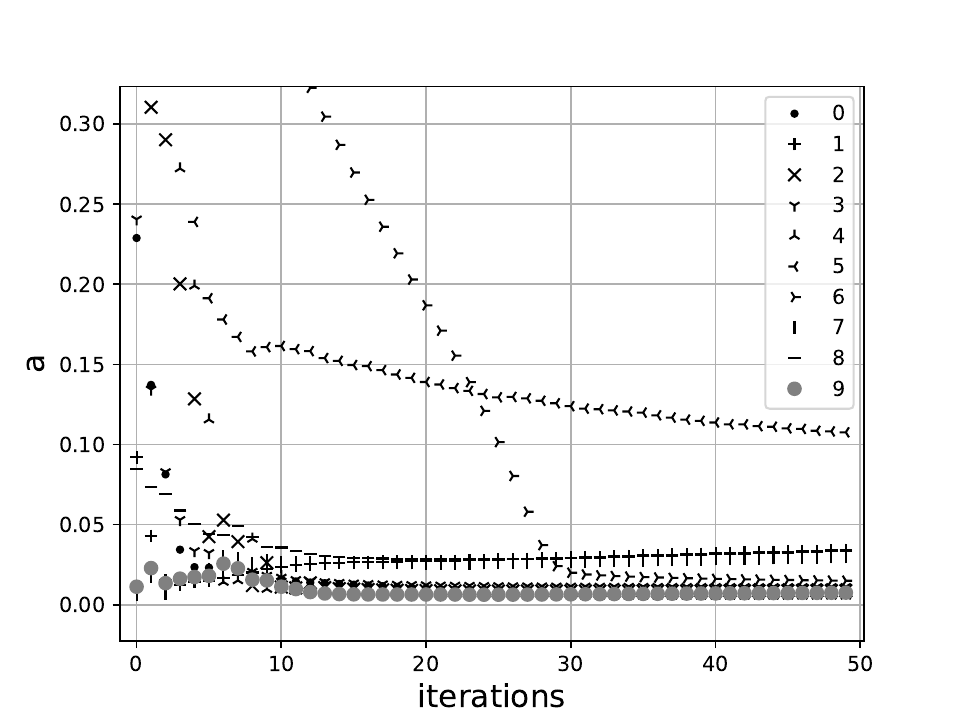}
\caption{$a$ values for each category (0th to 9th).}
 \label{fig7a}
\end{subfigure}
\begin{subfigure}[b]{0.45\textwidth}
\includegraphics[width=\textwidth]{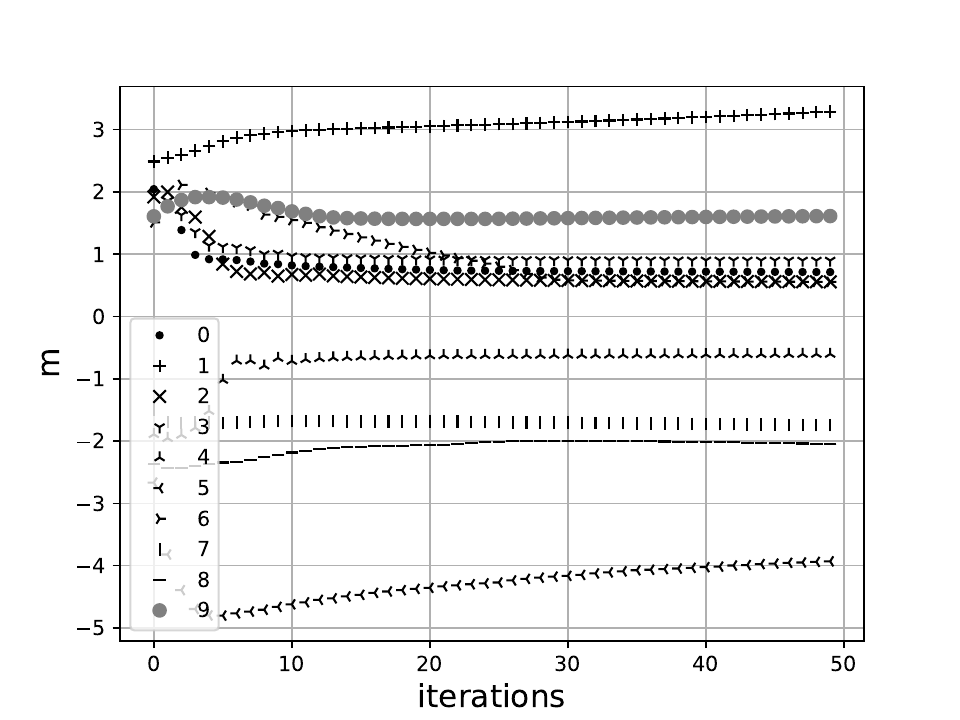}
\caption{Corresponding $m$ values for each category.}
 \label{fig7b}
\end{subfigure}
 \begin{subfigure}[b]{0.45\textwidth}
\includegraphics[width=\textwidth]{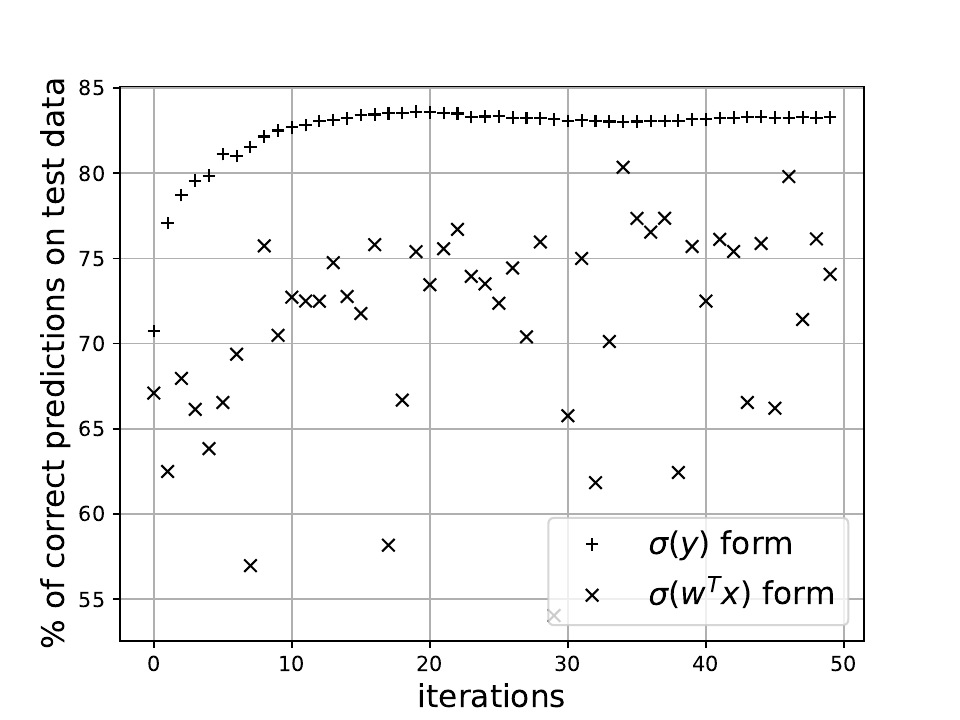}
\caption{Plots of accuracy of prediction for both forms of $\sigma$.}
 \label{fig7c}
\end{subfigure}
\begin{subfigure}[b]{0.45\textwidth}
\includegraphics[width=\textwidth]{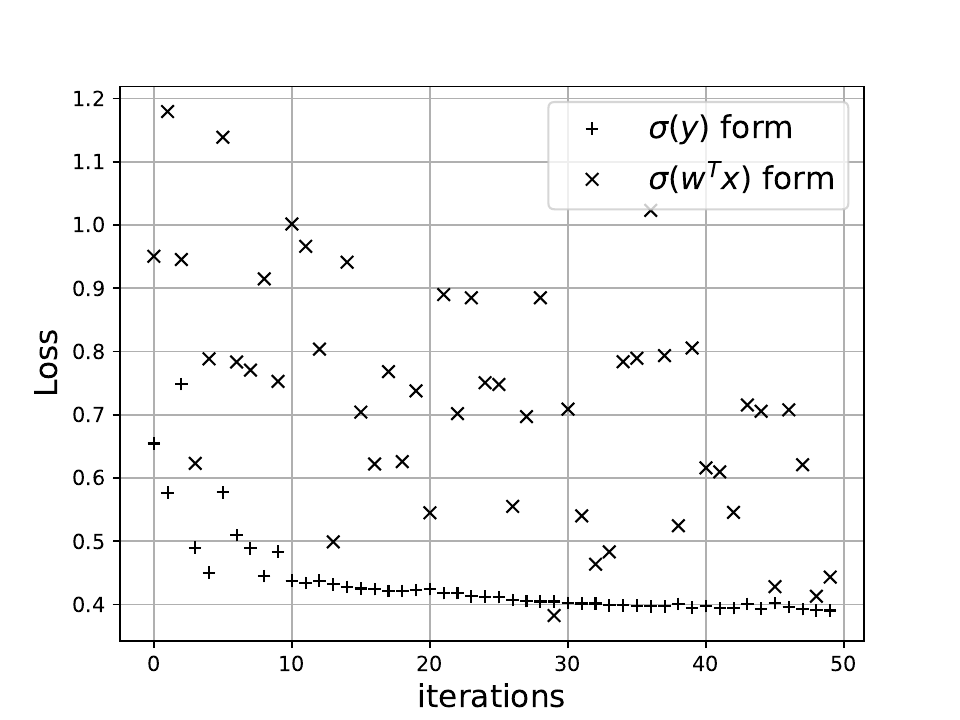}
\caption{Corresponding plots of loss function.}
 \label{fig7d}
\end{subfigure}
\caption{Estimation of parameters of continued fraction $a$ and $m$ are shown in (a) and (b). Accuracy and loss with both forms of $\sigma$ are compared in (c) and (d). Initial conditions: $a^{(0)}=1$ and $m^{(0)}=0$;  and $w_i^{(0)}$ are random values of a standard normal distribution. Final $a$ and $m$ values, $a^{(50)}$ and $m^{(50)}$ are shown in Table \ref{tab1}.}
\label{fig7}
\end{figure}

\begin{figure}[ht!]
\centering
 \begin{subfigure}[b]{0.45\textwidth}
\includegraphics[width=\textwidth]{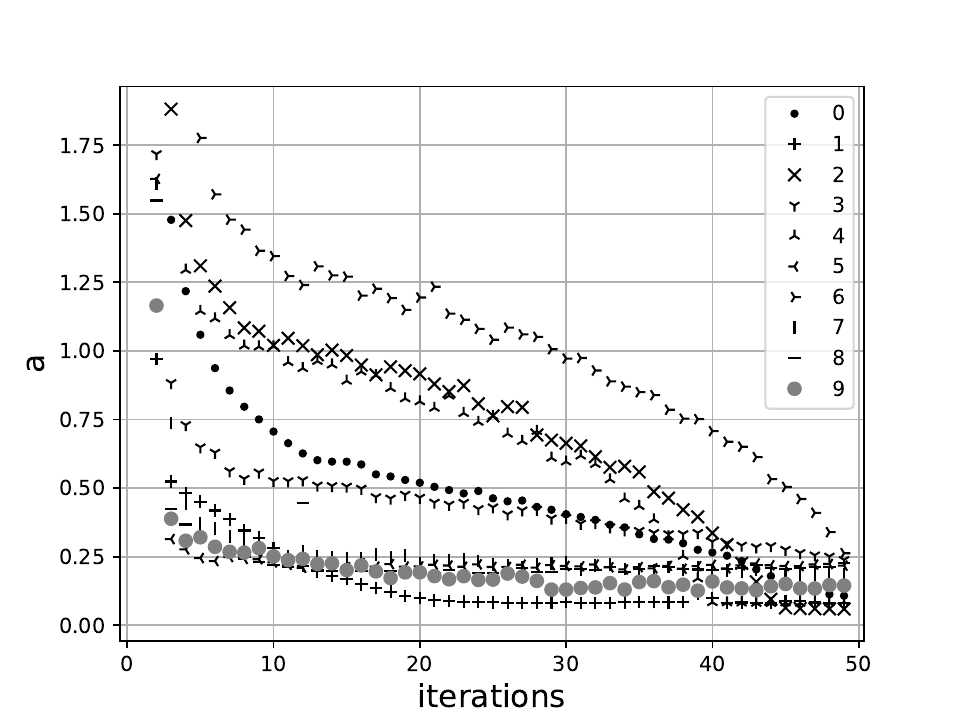}
\caption{$a$ values for each category (0th to 9th)}
\label{fig8a}
\end{subfigure}
\begin{subfigure}[b]{0.45\textwidth}
\includegraphics[width=\textwidth]{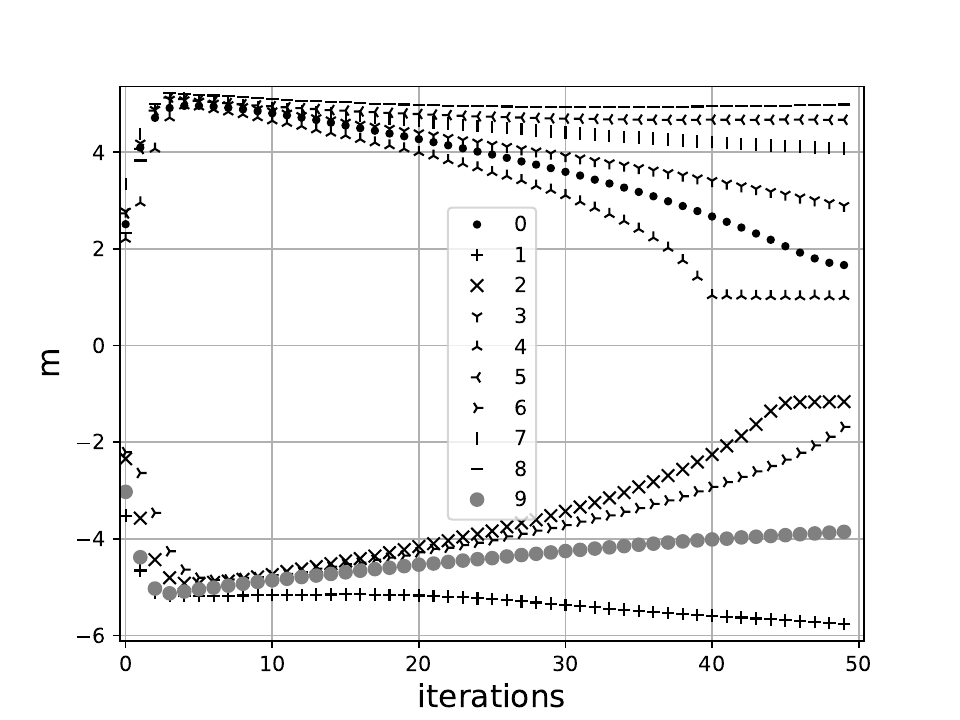}
\caption{$m$ values for each category (0th to 9th)}
\label{fig8b}
\end{subfigure}
 \begin{subfigure}[b]{0.45\textwidth}
\includegraphics[width=\textwidth]{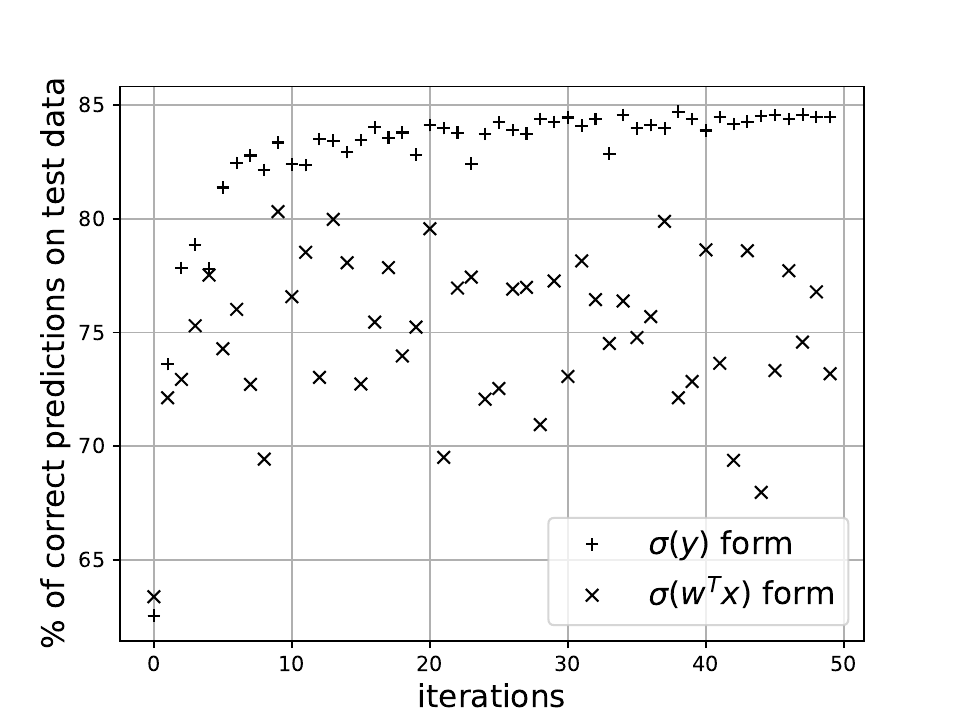}
\caption{Plots of accuracy of prediction for both forms of $\sigma$.}
\label{fig8c}
\end{subfigure}
\begin{subfigure}[b]{0.45\textwidth}
\includegraphics[width=\textwidth]{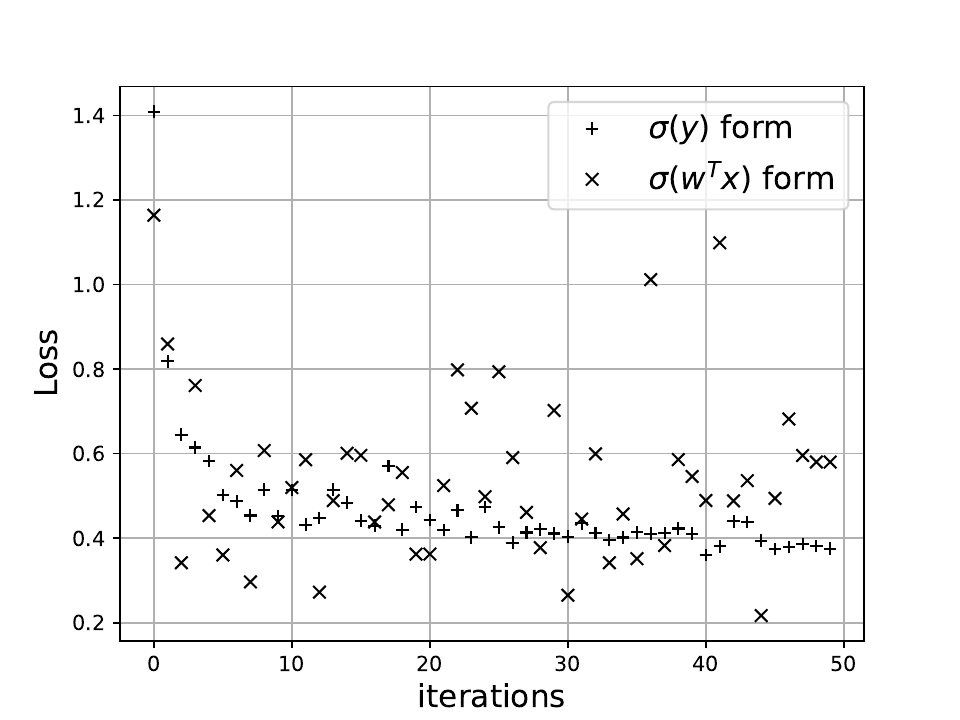}
\caption{Corresponding plots of loss function.}
\label{fig8d}
\end{subfigure}
\caption{Estimation of parameters of continued fraction $a$ and $m$ are shown in (a) and (b). Accuracy and loss with both forms of $\sigma$ are compared in (c) and (d). Initial conditions: $a^{(0)}=5$ and $m^{(0)}=0$;  and $w_i^{(0)}$ are random values of a standard normal distribution. Final $a$ and $m$ values, $a^{(50)}$ and $m^{(50)}$ are shown in Table \ref{tab1}.}
\label{fig8}
\end{figure}

\begin{figure}[ht!]
\centering
 \begin{subfigure}[t]{0.32\textwidth}
\includegraphics[width=\textwidth]{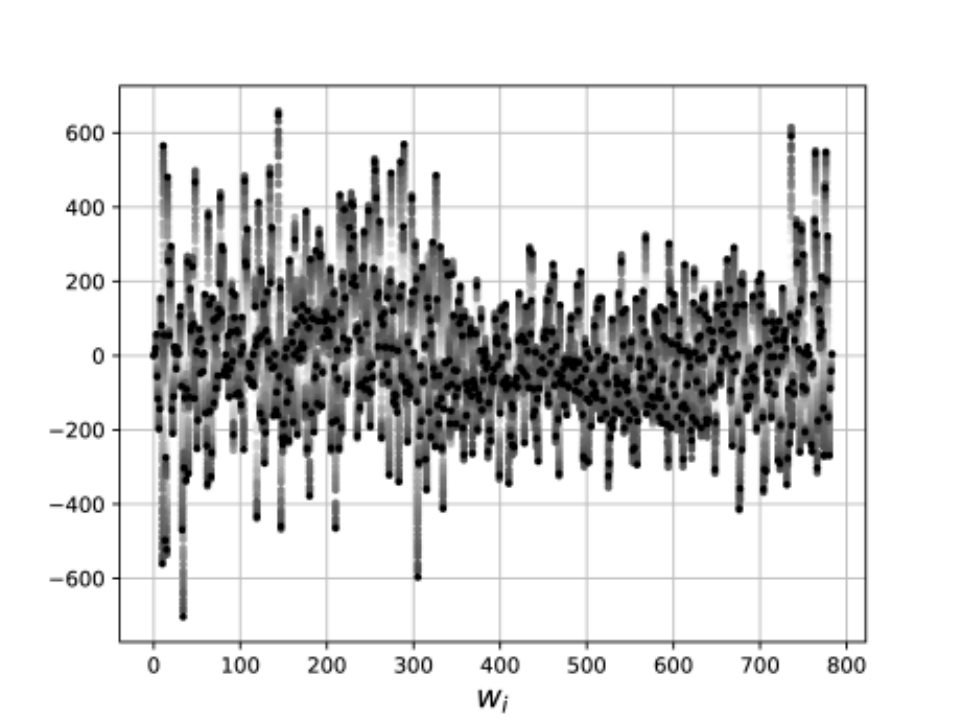}
\caption{$w_i'$s of $0^\text{th}$ class of $\sigma(w^T\textbf{x})$. $w_i^{(50)}\in(-700,700)$.}
\label{fig9a}
\end{subfigure}
\begin{subfigure}[t]{0.32\textwidth}
\includegraphics[width=\textwidth]{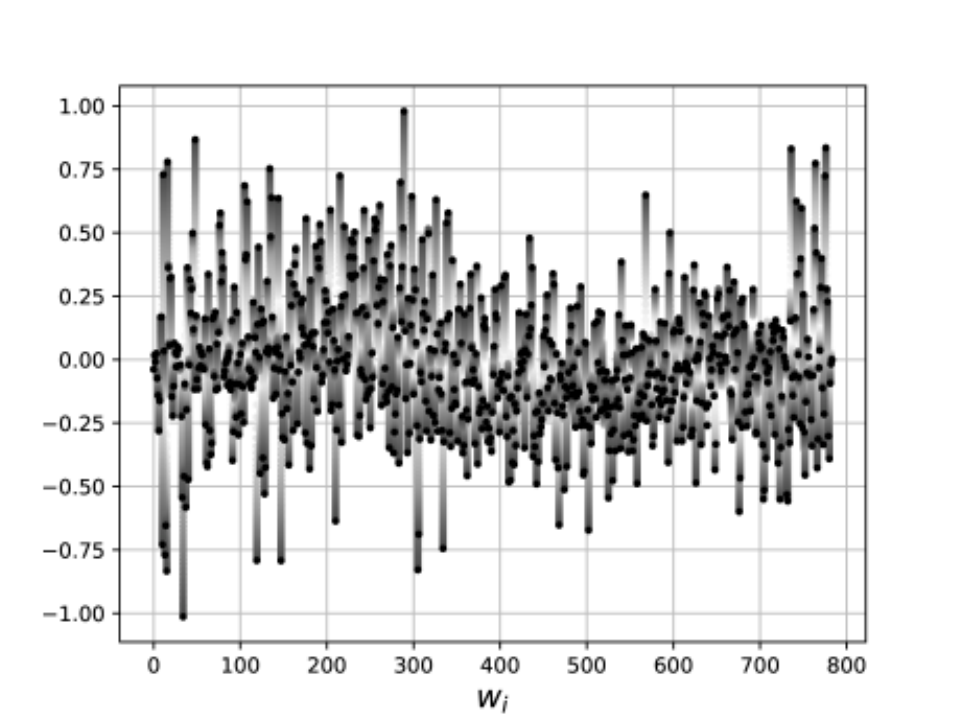}
\caption{$w_i'$s of $0^\text{th}$ class of $\sigma(y)$ with $a^{(0)}=1$, $m^{(50)}=0.71364981$. $w_i^{(50)}\in(-1.1,1)$}
\label{fig9b}
\end{subfigure}
\begin{subfigure}[t]{0.32\textwidth}
\includegraphics[width=\textwidth]{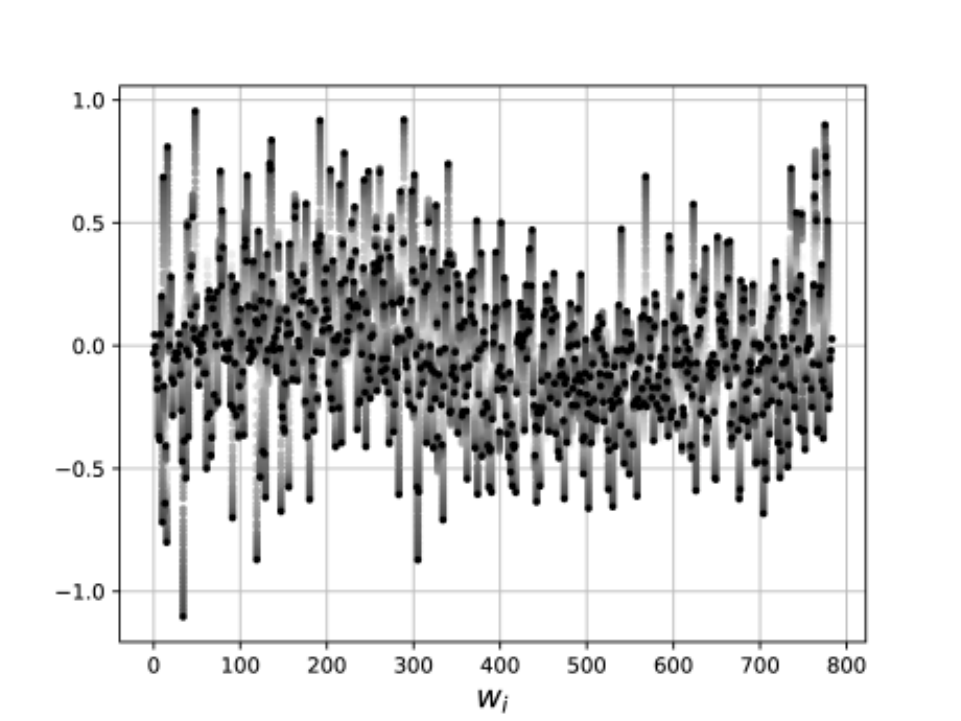}
\caption{$w_i'$s of $0^\text{th}$ class of $\sigma(y)$ with $a^{(0)}=5$, $m^{(50)}=1.66316775$.  $w_i^{(50)}\in(-1.5,1)$}
\label{fig9c}
\end{subfigure}
 \begin{subfigure}[t]{0.32\textwidth}
\includegraphics[width=\textwidth]{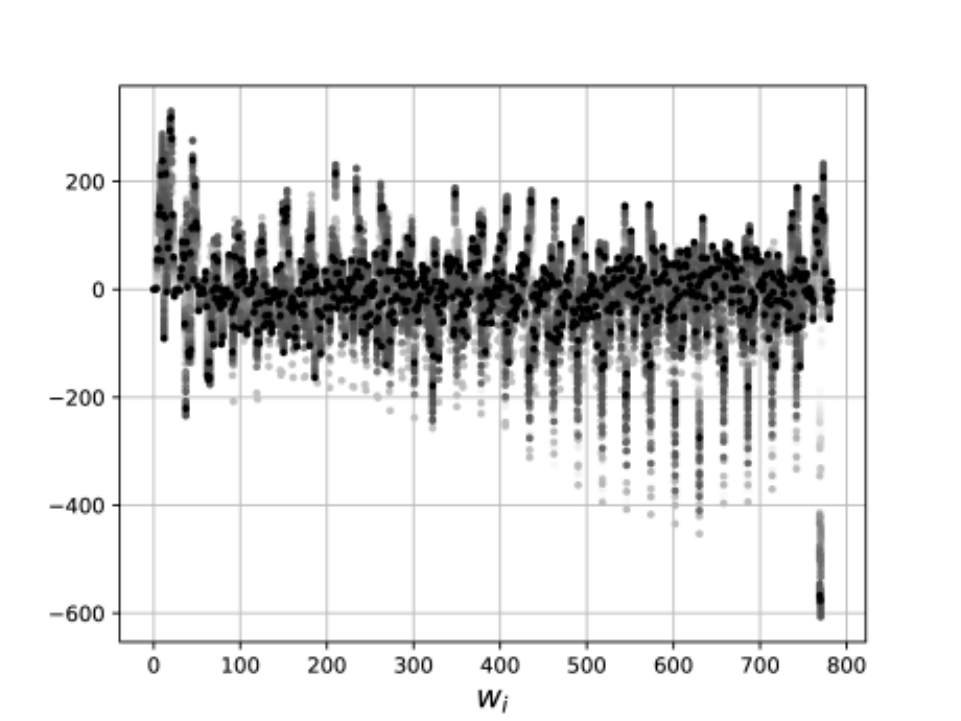}
\caption{$w_i'$s of $1^\text{st}$ class of $\sigma(w^T\textbf{x})$. $w_i^{(50)}\in(-650,400)$}
\label{fig9d}
\end{subfigure}
\begin{subfigure}[t]{0.32\textwidth}
\includegraphics[width=\textwidth]{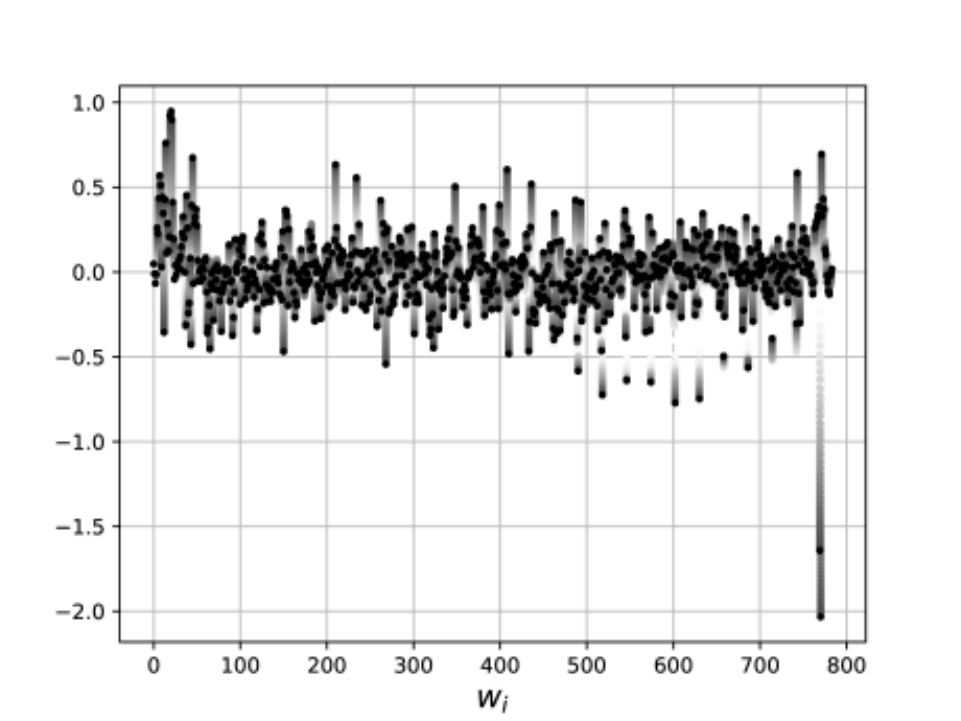}
\caption{$w_i'$s of $1^\text{st}$ class of $\sigma(y)$ with $a^{(0)}=1$, $m^{(50)}=3.28946245$.  $w_i^{(50)}\in(-2.2,1)$}
\label{fig9e}
\end{subfigure}
\begin{subfigure}[t]{0.32\textwidth}
\includegraphics[width=\textwidth]{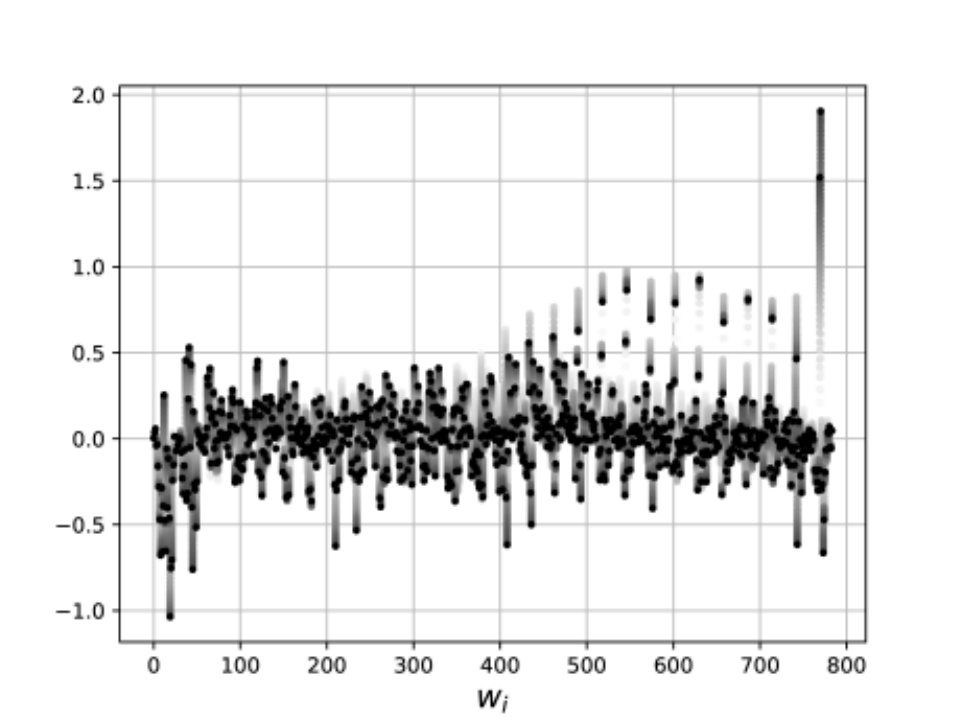}
\caption{$w_i'$s of $1^\text{st}$ class of $\sigma(y)$ with $a^{(0)}=5$, $m^{(50)}=-5.76584828$. $w_i^{(50)}\in(-1.2,2)$}
\label{fig9f}
\end{subfigure}
 \begin{subfigure}[t]{0.32\textwidth}
\includegraphics[width=\textwidth]{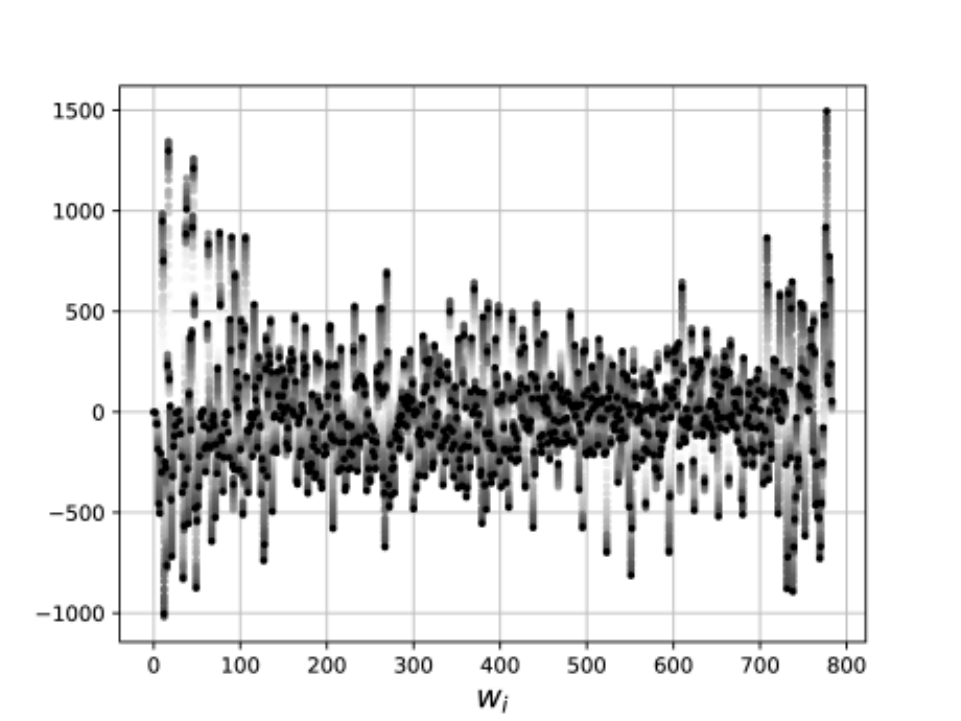}
\caption{$w_i'$s of $2^{\text{nd}}$ class of $\sigma(w^T\textbf{x})$. $w_i^{(50)}\in(-1100,1500)$}
\label{fig9g}
\end{subfigure}
\begin{subfigure}[t]{0.32\textwidth}
\includegraphics[width=\textwidth]{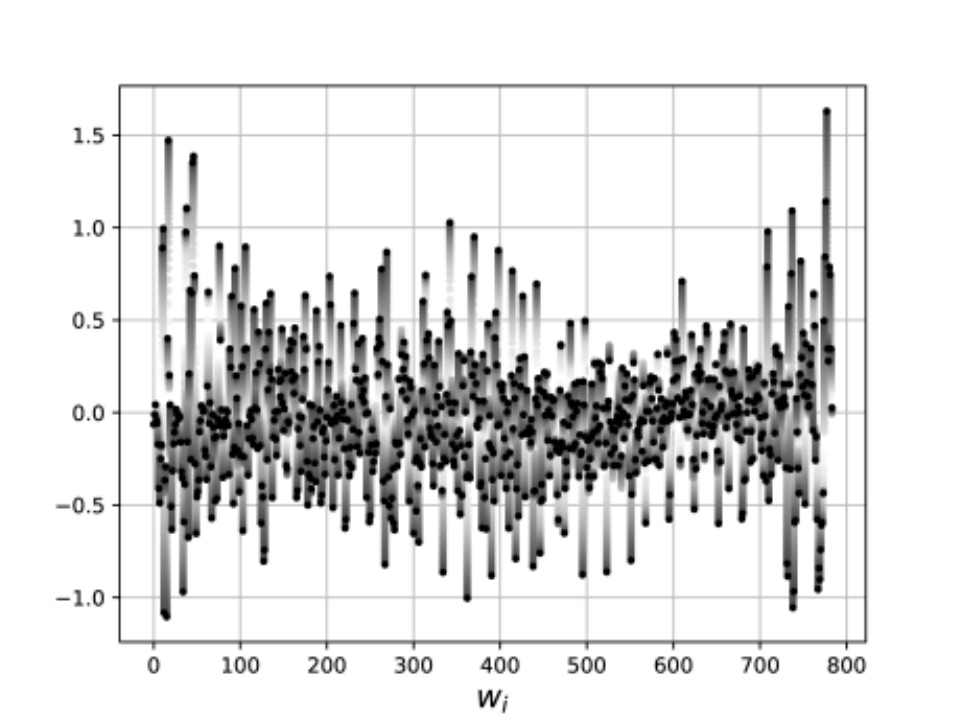}
\caption{$w_i'$s of $2^{\text{nd}}$ class of $\sigma(y)$ with $a^{(0)}=1$, $m^{(50)}=0.55567706$. $w_i^{(50)}\in(-1.5,2)$}
\label{fig9h}
\end{subfigure}
\begin{subfigure}[t]{0.32\textwidth}
\includegraphics[width=\textwidth]{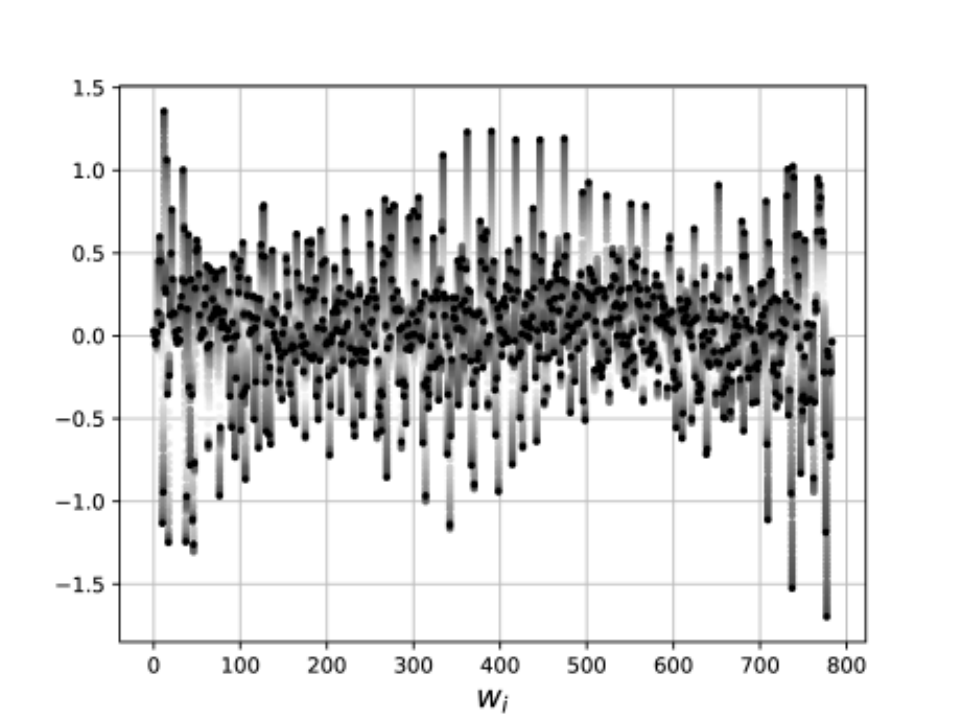}
\caption{$w_i'$s of $2^{\text{nd}}$ class of $\sigma(y)$ with $a^{(0)}=5$, $m^{(50)}=-1.16463021$. $w_i^{(50)}\in(-2,1.5)$}
\label{fig9i}
\end{subfigure}
\caption{$w_i'$s for $0^\text{th}$, $1^\text{st}$ and $2^\text{nd}$ categories of both $\sigma(w^T\textbf{x})$ and $\sigma(y)$, with $a^{(0)}=0$, $a^{(0)}=5$ are shown. \textbf{Boundedness:} The range of $w_i^{(50)}$ ($w_i'$s after 50 iterations) are shown above. (a), (d) and (g) have more variance. \textbf{Convergence:} The dark dots (`\drawbull[black]') are $w_i^{(50)}$s. 
The lighter shaded regions (`\drawlbulls[black]') are $w_i'$s from early iterations and
the darker regions (`\drawdbulls[black]') are $w_i'$s from later iterations (smooth convergence: `\drawbullsmooth[black]', oscillatory: `\drawbullosc[black]').}
\label{fig9}
\end{figure}

\paragraph{Implementation:}
The following are the steps taken to estimate the parameters:
\begin{enumerate}
\item Pixel values in greyscale images vary between 0 and 255. Dividing by 255, the pixel values are normalized  and $\textbf{x}\in[0,1]$ interval as shown in Fig. (\ref{fig6}).
\item The categories are represented as one-hot encoded vectors $p_{j}^*$ for the $j^{\text{th}}$ training sample.
\item Parameters $w_{0,1,2,3,...,784}$ are initialized as $w_i^{(0)}$, a $10\times (784+1)$ random matrix representing 784 pixel values of 10 categories and $w_0'$s for each category.
 $m$ is initialized as a 10x1 null vector with component $m^{(0)}=0$ for all categories.
\item $a$ is initialized as $a^{(0)}=1$ or $a^{(0)}=5$ for all categories. So, $a$ is a $10\times1$ vector.
\item 60,000 samples are divided into 100 batches of 600 samples and gradient descent based updates of parameters is carried out for each batch \cite{grad_desc}. This way of updating is known as mini-batch gradient descent (Sec. 2.3 of \cite{grad_desc}). Therefore, we have $n=600$ in Eqn. (\ref{eqloss}).
\item At each batch of $n$ samples:
\begin{enumerate}
\item We compute $y$ and update $a$ as 
\begin{equation}
a=a^{\text{(prev)}}-\frac{\alpha}{n}\sum_{j=1}^{n=600}(\sigma_j-p^*_j)\frac{-y_j^3}{1+3ay_j^2}.
\label{eqaup}
\end{equation}

\item If $a<0$ for any category, then we step back to $a=a^{\text{(prev)}}$ and reduce the step size as $\alpha=\alpha/1.1$ and reevaluate Eqn. (\ref{eqaup}). This is done till $a>0$ for all categories.
\item If $a>0$ for all categories, then we recompute $y$ using the updated $a$. $a$ and $y$ are then used to update $m$ using Eqn. (\ref{equpdates}) as
\begin{equation}
m = m^{\text{(prev)}}-\frac{1}{n}\sum_{j=1}^{n=600}(\sigma_j-p^*_j)\frac{(w^T\textbf{x})_j}{1+3ay_j^2}.
\label{eqmup}
\end{equation}
\item We again recompute $y$ using the updated $a$ and $m$. Now, using this $y$ and the updated $a$ and $m$ we update $w_i'$s using Eqn. (\ref{equpdates}) as
\begin{equation}
w_i = w_i^{\text{(prev)}} - \frac{1}{n}\sum_{j=1}^{n=600}(\sigma_j-p^*_j)\frac{mx_{ij}}{1+3ay_j^2}.
\label{eqwup}
\end{equation}
\item Substituting $m=1$ and $a=0$ in the above equation we get the updates for the $w_i'$s of $\sigma(w^T\textbf{x})$, which is the linear case. 
\end{enumerate}
\item The above steps are repeated for 100 batches completing one iteration over the entire 60,000 training samples.
\item After every iteration, the estimated parameters are used to classify the test image samples $\textbf{x}_{\text{test}}$ as $\sigma(y(w^T\textbf{x}_{\text{test}}))$ (or $\sigma(w^T\textbf{x}_{\text{test}})$) which are compared with the test outputs $p^*_{\text{test}}$ and the accuracy of classification is evaluated as the percentage of correct classifications.
\item After 50 iterations, $\sigma(y)$ form of accuracy and loss values are plotted and compared with the results of those obtained using $\sigma(w^T\textbf{x})$. These are shown in Fig. (\ref{fig7}) and (\ref{fig8}) for initial conditions of $a^{(0)}=1$ and $a^{(0)}=5$, respectively.
\end{enumerate}
The convergence of $a$ and $m$ values are also provided in Fig. (\ref{fig7}) and (\ref{fig8}). Their final values are provided in Table \ref{tab1}. The estimated $w_i'$s of both $\sigma(y)$ and $\sigma(w^T\textbf{x})$ are provided in Fig. (\ref{fig9}) and for initial conditions $a^{(0)}=1$ and $a^{(0)}=5$. The results are discussed in the following section.
\section{Discussions}
\subsection{Boundedness}
In Figs. (\ref{fig1}) and (\ref{fig2}), $y$ values on $xy-$ plane and $\theta$ values on $(r,\theta)-$ plane are bounded by the linear version $y=mx$ or $\tan\theta=m$, respectively. This can also be seen from Fig. (\ref{fig5}). Even when the terms of truncated continued fraction (such as $C/\mathcal{C}$, $G/\mathcal{G}$) diverge from i+ii curves, they are all bounded by $y=x$ in Fig. (\ref{fig5}). 

The bounding nature can also be seen in Fig. (\ref{fig9}). The variance of $w_i'$s is considerably decreased due to the bounding nature of continued fraction of straight lines. Yet, the variations in $w_i'$s are captured and the accuracy of classification has been enhanced. In Fig. (\ref{fig9a}), (\ref{fig9d}) and (\ref{fig9g}) the values of $w_i'$s vary between (-700,700), (-650,400) and (-1000,1500) intervals, respectively. These are $w_i'$s of $\sigma(w^T\textbf{x})$. Corresponding $w_i'$s of $\sigma(y)$ in Fig. (\ref{fig9}) vary in the range (-2,2) for both $a^{(0)}=1$ and $a^{(0)}=5$. High values of $w_i'$s make the model more sensitive to noise and also become less reliable due to overfitting of training data. This is also the reason for more fluctuations, across iterations, in loss and accuracy using $\sigma(w^T\textbf{x})$ compared to those using $\sigma(y)$ as shown in Fig. (\ref{fig8c}), (\ref{fig8d}), (\ref{fig9c}) and (\ref{fig9d}).
\subsection{Convergence}
From Fig. (\ref{fig7a}), (\ref{fig7b}) and Fig. (\ref{fig8a}), (\ref{fig8b}) it can be seen that $a$ and $m$ converge to a constant value for each category. Once $a$ and $m$ reach almost constant values the plots of loss and accuracy become smooth. For example, for $a^{(0)}=5$, the plots of Fig. \ref{fig8} become smoother after 40 iterations, after most of $a$ and $m$ values converge. The plots of accuracy and loss of Fig. \ref{fig7} for $a^{(0)}=1$ are much smoother than those of $a^{(0)}=5$. For $a^{(0)}=1$, $a$ and $m$ converge sooner than those with $a^{(0)}=5$. 

Also, $a$ and $m$ values converge simultaneously towards constant values in Fig. (\ref{fig7a}), (\ref{fig7b}), (\ref{fig8a}) and (\ref{fig8b}). In Fig. (\ref{fig7a}) and (\ref{fig7b}), for the $6^{\text{th}}$ category (shown as `$\Yright$' ), $a$ and $m$ converge at the $29^{\text{th}}$ iteration. And for the $9^{\text{th}}$ category (shown as `$\tikz\draw[black!45!white,fill=black!45!white] (0,0) circle (.5ex);$') $a$ and $m$ converge at the $12^{\text{th}}$ iteration. Similarly, in Fig. (\ref{fig8a}) and (\ref{fig8b}), for the $2^{\text{nd}}$ category (shown as `$\times$'), $a$ and $m$ converge at the $45^{\text{th}}$ iteration. While for the $4^{\text{th}}$ (shown as `$\Yup$') $a$ and $m$ converge at the $40^{\text{th}}$ iteration. $a$ values for $1^{\text{st}}$ (shown as `+') and $9^{\text{th}}$ fluctuates mildly as in Fig. (\ref{fig8a}), corresponding $m$ values in Fig. (\ref{fig8b}) vary slowly.

Smooth convergence of estimated $w_i'$s of $\sigma(y)$ can also be seen in Fig. (\ref{fig9}). In Fig. (\ref{fig9}) regions of lighter greyscale areas `\drawlbulls'
of each $w_i$ represent values of $w_i$ from the earlier iterations and the darker areas `\drawdbulls' are from later iterations 
and the dark dot `\drawbull' represents the final value of a $w_i$ after 50 iterations. Values at the later stages of iterations occupy the darker region.
$w_i'$s of Fig. (\ref{fig9b}), (\ref{fig9c}), (\ref{fig9e}), (\ref{fig9f}), (\ref{fig9h}) and (\ref{fig9i}) converge almost monotonically towards the final values, i.e., the dark dot for each $w_i$ is at the `edge' as a final value for each $w_i$, as in: `\drawbullsmooth'
Whereas those in Fig. (\ref{fig9a}), (\ref{fig9d}) and (\ref{fig9g}) are more oscillatory i.e., the dark dot moves back into the grey shade region, as in: `\drawbullosc'. 
Therefore, other than boundedness, a smoother convergence is also achieved using $y$.
\subsection{$xy-$plane representation}
\begin{figure}[ht!]
\centering
 \begin{subfigure}[b]{0.45\textwidth}
\includegraphics[width=\textwidth]{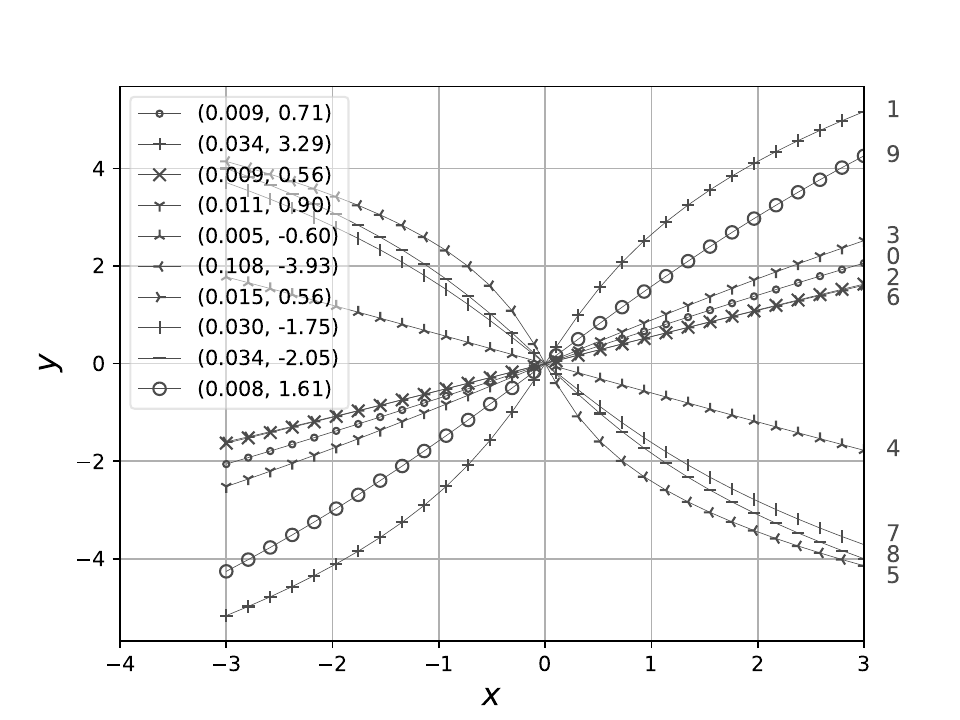}
\caption{$a^{(0)}=1,\ m^{(0)}=0$ and $w_i'$s are randomly initialized.}
\label{fig10a}
\end{subfigure}
\begin{subfigure}[b]{0.45\textwidth}
\includegraphics[width=\textwidth]{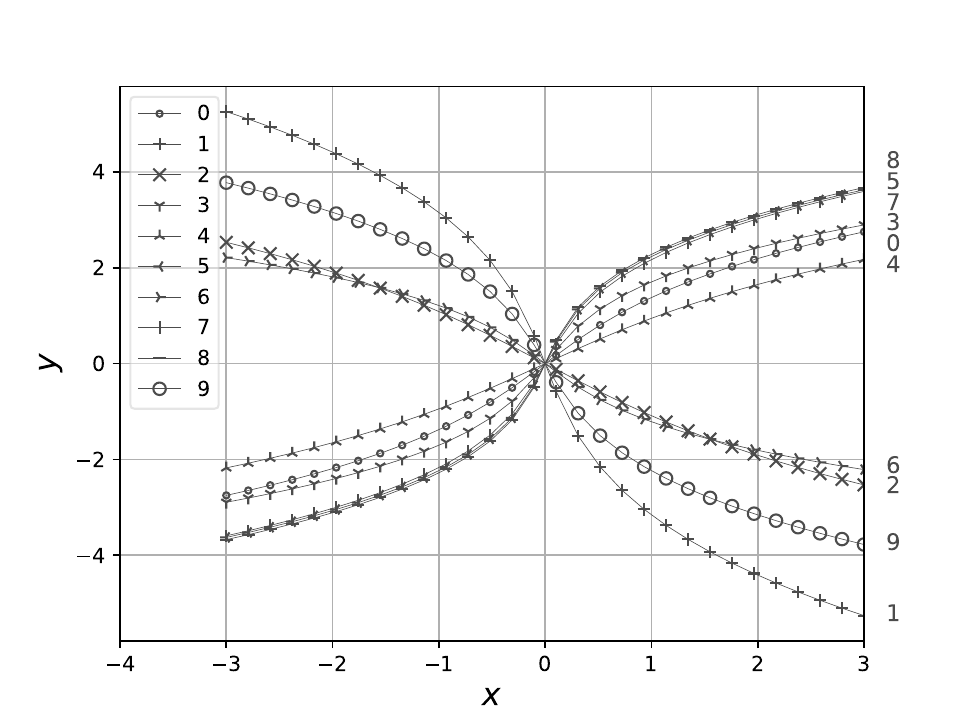}
\caption{$a^{(0)}=5,\ m^{(0)}=0$ and $w_i'$s are randomly initialized.}
\label{fig10b}
\end{subfigure}
 \begin{subfigure}[b]{0.45\textwidth}
\includegraphics[width=\textwidth]{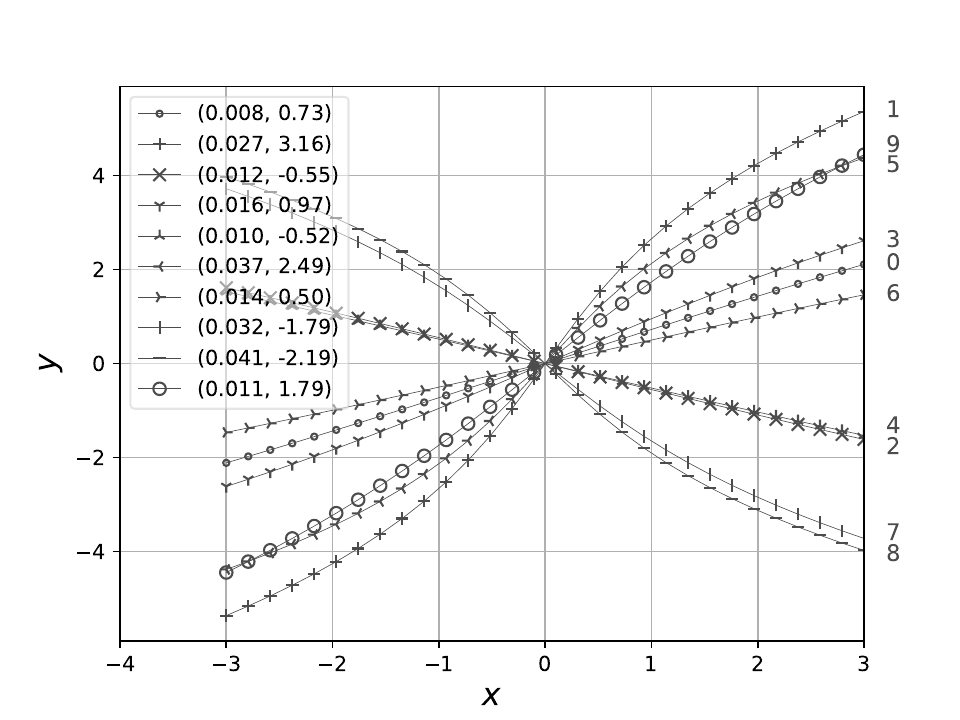}
\caption{$a^{(0)}=1,\ m^{(0)}=0$ and $w_i'$s are randomly initialized (different from above).}
\label{fig10c}
\end{subfigure}
\begin{subfigure}[b]{0.45\textwidth}
\includegraphics[width=\textwidth]{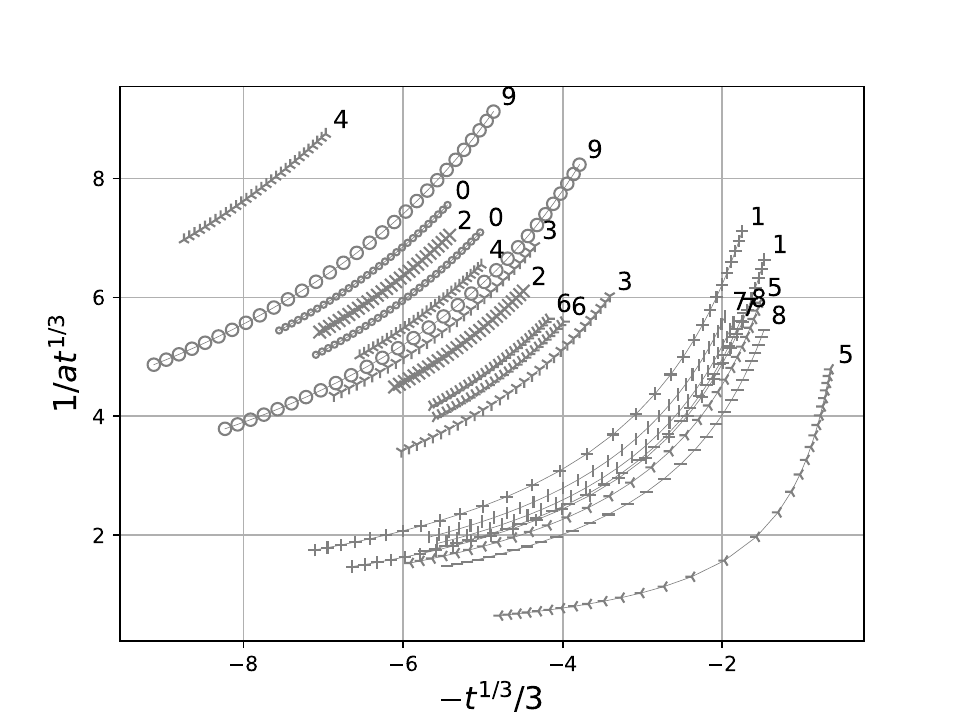}
\caption{Plots of (a) and (c) on i,ii- plane with $a^{(0)}=1,x\in[-3,3]$.}
\label{fig10d}
\end{subfigure}
\caption{For various initial conditions, the final values of $a$ and $m$ are shown as coordinates ($a^{(50)},m^{(50)}$) for each category in (a) and (c). (d) shows the plots of (a) and (c) on i,ii- plane.}
\label{fig10}
\end{figure}

In Fig. (\ref{fig9e}) and (\ref{fig9f}), the variations of $w_i'$s seem inverted for $a^{(0)}=1$ and $a^{(0)}=5$. This is because the final $m$ values are of opposite sign for $a^{(0)}=1$ and $a^{(0)}=5$ given by $m^{(50)}=3.28946245$ and -5.76584828, respectively for the $1^{\text{st}}$ category. This is also the case for $w_i'$s in Fig. (\ref{fig9h}) and (\ref{fig9i}) for the $2^{\text{nd}}$ category. Thus, $m$ acts as a common scaling for all $w_i'$s of each category.
With values of $a$ and $m$ obtained after 50 iterations, we can plot $y$ (or i+ii curves) for each category on $xy-$ plane. Although this is a multi-dimensional problem, $xy-$ plots are possible because $a$ and $m$ converge to unique values for each category. The plots are shown in Fig. (\ref{fig10}) for various initial conditions. 

Plots on i-ii plane (as in Fig. (\ref{fig4b})) is also shown in Fig. (\ref{fig10d}). Plots in Fig. (\ref{fig10d}) help us to observe the categorical models that are more sensitive to the changes in initial conditions. Thus, plots on i-ii plane help us to identify sensitive patterns within the classification model. 
\begin{table}[h!]
\centering
\begin{tabular}{|c|c|c|c|c|}
\hline
\multirow{3}{*}{Categories} &\multicolumn{2}{|c|}{Initial condition $a^{(0)}=1$, $m^{(0)}=0$}& \multicolumn{2}{|c|}{Initial condition $a^{(0)}=5$, $m^{(0)}=0$}\\\cline{2-5}
&\multicolumn{4}{|c|}{After 50 iterations}\\
\cline{2-5} & $a^{(50)}$&$m^{(50)}$&$a^{(50)}$&$m^{(50)}$
\\\hline
0&0.00933464&0.71364981&0.10750743&1.66316775 \\\hline
1&0.03415196&3.28946245&0.08268326&-5.76584828\\\hline
2&0.00873391&0.55567706&0.05923945&-1.16463021 \\\hline
3&0.01116299&0.89931317&0.23828527&2.89171539\\\hline
4&0.00546778&-0.60251737&0.08513255&1.01813209 \\\hline
5&0.10757327&-3.93066244&0.21727861&4.67005625\\\hline
6&0.01494351&0.5553683& 0.26259974&-1.68538874 \\\hline
7&0.02986627&-1.74514672&0.18564456& 4.07392969\\\hline
8&0.03357115&-2.04798275&0.22733055& 4.98591511 \\\hline
9&0.00750415& 1.61222652&0.14469838& -3.8574718\\\hline
\end{tabular}
\caption{Final values of $a$ and $m$ after 50 iterations for $a^{(0)}=1$ and $a^{(0)}=5$.}
\label{tab1}
\end{table}

\section{Conclusions}
Real-valued continued fraction of straight lines, in this work, converges to the real solution of a cubic equation in $y$. The nonlinear and the input parts of the equation are parameterized with the parameters $a$ and $m$, respectively. Parameters estimated using the real-valued continued fraction of linear scale has less variance than those estimated with linear scale. Moreover, with the two parameters $a$ and $m$, the convergence of parameters of regression is almost monotonic and the step size is adaptive and bounded. Thus, the usefulness of the bounded behavior of real-valued continued fraction has been demonstrated with the image classification problem. We have also represented a multi-dimensional problem on planar plots: $xy-$ plane or i,ii- plane using $a$ and $m$. To conclude, $a$ and $m$ form a bounded nonlinear co-ordinate system with straight lines at its asymptotic limit $a\rightarrow 0$. Each coordinate $(a,m)$ is a curve on the $xy-$ plane. Once $a$ is fixed, $xy-$ plane can be parametrized with $m$ in a bounded manner.

\bibliographystyle{unsrt}

\begin{thebibliography}{00}
\bibliographystyle{unsrt}

\bibitem{euler} Euler, L. (2004) On the Transformation of Infinite Series to Continued Fractions (D. W. File, Trans.). Reading Classics: Euler.
\url{https://people.math.osu.edu/sinnott.1/ReadingClassics/continuedfractions.pdf} (Original work published 1785).

\bibitem{method_book} Abu-Mostafa, Y.S., Magdon-Ismail, M., and Lin, H.T. (2012). Learning from data (Vol. 4, p. 4). New York: AMLBook.

\bibitem{grad_desc} Ruder, S. (2017). An overview of gradient descent optimization algorithms. arXiv preprint arXiv:1609.04747.
 
\bibitem{f_mnist} Xiao, H., Rasul, K., Vollgraf, R. (2017). Fashion-mnist: a novel image dataset for benchmarking machine learning algorithms. arXiv preprint arXiv:1708.07747.

\end{thebibliography}

\end{document}